\pdfoutput=1

\documentclass[11pt]{article}

\usepackage[preprint]{acl}

\usepackage{times}
\usepackage{latexsym}

\usepackage[T1]{fontenc}

\usepackage[utf8]{inputenc}

\usepackage{microtype}

\usepackage{inconsolata}

\usepackage{graphicx}
\usepackage{booktabs}
\usepackage{hyperref}
\usepackage{url}

\usepackage{hyperref}
\usepackage{url}
\usepackage{hyperref}
\usepackage[most]{tcolorbox}
\usepackage{wrapfig}
\usepackage{graphicx,xcolor,float}
\usepackage{subcaption}
\usepackage{threeparttable}
\usepackage{algorithm}
\usepackage{pifont}
\usepackage[noend]{algorithmic}
\usepackage{colortbl}
\usepackage{color}
\usepackage{multirow}
\usepackage{tabularx}
\usepackage{float}
\usepackage{graphicx}
\usepackage{booktabs}
\usepackage{arydshln}
\usepackage{enumitem}
\usepackage{wrapfig}
\usepackage{caption}
\usepackage{graphicx}
\usepackage{makecell}
\usepackage{float} 
\usepackage{makecell}
\usepackage{tabularx}
\usepackage{twemojis}
\usepackage{amssymb,mathrsfs,amsmath}

\usepackage{pifont}
\usepackage[export]{adjustbox}
\usepackage{xcolor}
\usepackage{xspace}
\usepackage{shadowtext}
\usepackage{anyfontsize}
\usepackage{array}

\hypersetup{
    colorlinks=true,
    linkcolor=red,
    citecolor=cyan,
    filecolor=magenta,      
    urlcolor=magenta,
    }

\definecolor{bittersweet}{rgb}{1.0, 0.44, 0.37}
\definecolor{mygreen}{rgb}{0.29, 0.7, 0.48}
\definecolor{my_green}{RGB}{51,102,0}
\definecolor{my_yellow}{RGB}{255,165,0}
\definecolor{my_red}{RGB}{204, 0, 0}

\usepackage{pifont}
\definecolor{demphcolor}{RGB}{144,144,144}

\definecolor{mygray}{gray}{0.4}
\hypersetup{
    colorlinks=true,
    linkcolor=red,
    citecolor=cyan,
    filecolor=magenta,      
    urlcolor=magenta,
    }
\usepackage{xcolor} 
\usepackage{xcolor}

\usepackage[font=small,labelfont=bf]{caption}  
\usepackage{makecell}
\usepackage{tabulary}
\definecolor{ada_green}{rgb}{0,205,205}
\definecolor{glt_red}{rgb}{109,205,255}

\definecolor{backred}{RGB}{255, 190, 190}
\definecolor{backblue}{RGB}{210, 230, 250}
\definecolor{backgrey}{RGB}{220, 220, 220}
\newcommand{\high}{\cellcolor{backblue}}

\newcommand{\best}{\cellcolor{backred}}

\definecolor{shadecolor}{RGB}{237,237,237}

\newcommand{\dataset}{\textsc{EssayJudge}\xspace}

\title{EssayJudge: A Multi-Granular Benchmark for Assessing Automated Essay Scoring Capabilities of Multimodal Large Language Models}

\author{Jiamin Su\textsuperscript{\rm 1,\rm 2,*}, Yibo Yan\textsuperscript{\rm 1,\rm 3,*}, Fangteng Fu\textsuperscript{\rm 1}, Han Zhang\textsuperscript{\rm 1}, \\
\textbf{Jingheng Ye}\textsuperscript{\rm 4},
\textbf{Xiang Liu}\textsuperscript{\rm 1,\rm 3},
\textbf{Jiahao Huo}\textsuperscript{\rm 1},
\textbf{Huiyu Zhou}\textsuperscript{\rm 5},
\textbf{Xuming Hu}\textsuperscript{\rm 1,\rm 2,\rm 3,}\footnotemark[2]\\
\fontsize{9.0pt}{\baselineskip}\selectfont \textsuperscript{\rm 1}The Hong Kong University of Science and Technology (Guangzhou)\\
\fontsize{9.0pt}{\baselineskip}\selectfont \textsuperscript{\rm 2}Beijing Future Brain Education Technology Co., Ltd.\\
\fontsize{9.0pt}{\baselineskip}\selectfont \textsuperscript{\rm 3}The Hong Kong University of Science and Technology, 
\fontsize{9.0pt}{\baselineskip}\selectfont \textsuperscript{\rm 4}Tsinghua University\\
\fontsize{9.0pt}{\baselineskip}\selectfont \textsuperscript{\rm 5}Guangxi Zhuang Autonomous Region Big Data Research Institute\\
\fontsize{9.0pt}{\baselineskip}\selectfont \texttt{jsu360@connect.hkust-gz.edu.cn}, \texttt{yanyibo70@gmail.com},
\texttt{xuminghu@hkust-gz.edu.cn}
}

\begin{document}
\maketitle
\renewcommand{\thefootnote}{\fnsymbol{footnote}}
\footnotetext[1]{Co-first authors with equal contribution.}
\footnotetext[2]{Corresponding author.}
\renewcommand{\thefootnote}{\arabic{footnote}}
\begin{abstract}
Automated Essay Scoring (AES) plays a crucial role in educational assessment by providing scalable and consistent evaluations of writing tasks. However, traditional AES systems face three major challenges: \ding{182} reliance on handcrafted features that limit generalizability, \ding{183} difficulty in capturing fine-grained traits like coherence and argumentation, and \ding{184} inability to handle multimodal contexts. In the era of Multimodal Large Language Models (MLLMs), we propose \dataset, the \textbf{first multimodal benchmark to evaluate AES capabilities across lexical-, sentence-, and discourse-level traits}. By leveraging MLLMs' strengths in trait-specific scoring and multimodal context understanding, \dataset aims to offer precise, context-rich evaluations without manual feature engineering, addressing longstanding AES limitations. Our experiments with 18 representative MLLMs reveal gaps in AES performance compared to human evaluation, particularly in discourse-level traits, highlighting the need for further advancements in MLLM-based AES research. The dataset and code are available at \href{https://github.com/jsu360/EssayJudge}{this URL}.
\end{abstract}

\section{Introduction}
Automated Essay Scoring (AES) has become an essential tool in educational assessment, providing efficient and consistent scoring for large-scale writing tasks \cite{ye2025position,ramesh2022automated,li2024applying,wu2024unveiling,xia2024empirical}. While AES systems have significantly reduced the workload of human graders, they still \textit{face challenges in delivering accurate and detailed evaluations, particularly for trait-specific scoring}, which assesses individual aspects of writing quality, such as coherence, creativity, and argumentation \cite{song2024automated,pack2024large,ruseti2024automated}. Such detailed feedback is critical for guiding students in improving their writing skills, but remains difficult to achieve with existing methods.

\begin{figure}[!t]
    \centering
    \includegraphics[width=\linewidth,scale=1.00]{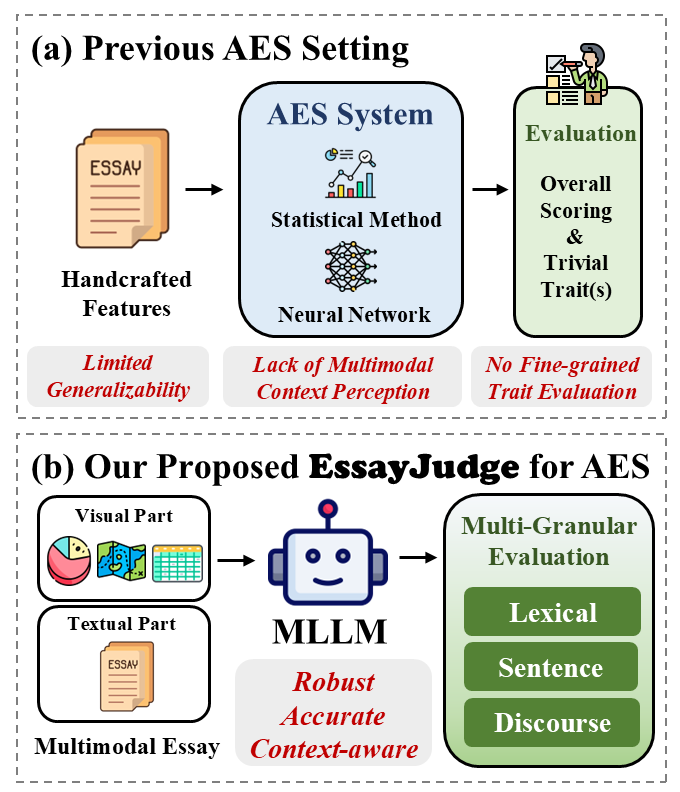}
    \caption{Comparison of task settings between the previous evaluation paradigm (a) and our proposed \dataset benchmark (b) on automated essay scoring task.}
    \label{fig:paradigm_comparison}
    \vspace{-6mm}
\end{figure}

Traditional AES approaches, including statistical models such as Support Vector Machines, rely heavily on handcrafted features such as word frequency and essay length \cite{yang2024unveiling,jansen2024individualizing,uto2020neural}. As illustrated in Figure \ref{fig:paradigm_comparison} (a), they often suffer from \ding{182} relying on manually engineered features, thus limiting the generalizability across diverse data; \ding{183} failing to model fine-grained traits such as logical structure and argument persuasiveness; \ding{184} inability to handle multimodal context, thus struggling to deliver comprehensive and context-aware evaluations \cite{lim2021comprehensive,uto2021review,wang2022bert}.

The emergence of Large Language Models (LLMs) and Multimodal Large Language Models (MLLMs) offers a promising solution to these challenges \cite{xiao2024automation,mansour2024fewshoting,ding2024automated,luo2025llm4sr}. Unlike traditional models, LLMs can capture rich semantic representations directly from text, allowing them to evaluate essays holistically and provide detailed feedback on specific traits \cite{gao2024automatic,maity2024future}. Furthermore, MLLMs extend this capability by integrating text and image inputs, enabling a deeper understanding of multimodal essay context \cite{lee2023multimodality,xu2024foundation}. This not only improves scoring accuracy but also addresses the need for nuanced evaluations in more complex writing scenarios.

Therefore, we propose \dataset, the \textbf{first multimodal benchmark for assessing the multi-granular AES capabilities of MLLMs}. As shown in Figure \ref{fig:paradigm_comparison} (b), \dataset aims to address the aforementioned gaps by \ding{182} refraining from manually engineered features, using MLLMs’ inherent capabilities to automatically capture complex linguistic patterns and contextual cues, which allows for better generalization across diverse datasets; \ding{183} leveraging MLLMs’ ability to model multi-granular traits (\textit{i.e.,} lexical-, sentence-, and discourse-levels), enabling more precise and nuanced trait-specific evaluations; \ding{184} incorporating multimodal inputs including text and image components, enabling MLLM to handle complex context, which is crucial for essays on intricate topics.

Through extensive experiments, we evaluated 18 representative open-source and closed-source MLLMs, yielding the following key insights: (i) open-source MLLMs generally perform poorly in AES compared to closed-source MLLMs, particularly GPT-4o; (ii) closed-source MLLMs tend to assign lower scores across multiple traits compared to human evaluators, reflecting their stricter scoring criteria; (iii) closed-source MLLMs perform better in evaluating essays based on single-image setting compared to multi-image one. In general, our findings highlight that there is still a noticeable gap in AES performance compared to human evaluators, particularly in discourse-level traits, underscoring the necessity for further LLM research.

Our contributions can be summarized as follows:
\begin{itemize}[leftmargin=*]
    \item We introduce the \textbf{first multimodal AES dataset} \dataset, comprising over 1,000 high-quality multimodal English essays, each of which has undergone rigorous multi-round human annotation and verification.
    \item We propose a \textbf{trait-specific scoring framework} that enables comprehensive evaluation with ten multi-granular metrics, covering three dimensions: lexical, sentence, and discourse levels. 
    \item We conduct an \textbf{in-depth evaluation of 18 state-of-the-art MLLMs and human evaluation} on \dataset, using Quadratic Weighted Kappa (QWK) as the primary metric to assess the trait-specific scoring performance.
\end{itemize}
By bridging the gaps in existing AES benchmarks, \dataset contributes to the development of more accurate, robust, and context-aware MLLM-based essay scoring systems in the era of AGI.

\section{Related Work}

\begin{figure}[t]
\centering
\begin{minipage}{\columnwidth} 
    \centering 
    \small
    \renewcommand\tabcolsep{2pt} 
    \renewcommand\arraystretch{1.1}
    \resizebox{\columnwidth}{!}{
        \begin{tabular}{lccccc}
            \toprule
            \textbf{Benchmarks} & \textbf{Venue} & \textbf{Size} & \textbf{\#Topics} & \textbf{Modality} & \textbf{\#Traits} \\
            \midrule
            $\text{ASAP}_{\text{AES}}$ \cite{cozma2018ASAP} & ACL & 17,450 & 8 & T & 0 \\
            ASAP++ \cite{mathias2018asap++} & ACL & 10,696 & 6 & T & 8 \\
            CLC-FCE \cite{yannakoudakis2011CLCFCE} & ACL & 1,244 & 10 & T & 0 \\
            TOEFL11 \cite{lee2024TOEFL11} & EMNLP & 1,100 & 8 & T & 0 \\
            ICLE \cite{granger2009icle} & COLING & 3,663 & 48 & T & 4\\
            AAE \cite{stab2014AAE} & COLING & 102 & 101 & T & 1 \\
            ICLE++ \cite{li2024icle++} & NAACL & 1,008 & 10 & T & 10 \\
            CREE \cite{bailey2008CREE} & BEA & 566 & 75 & T & 1 \\
            \midrule
            \dataset (Ours) & - & 1054 & \colorbox{red!20}{125} & T,\colorbox{red!20}{I} & \colorbox{red!20}{10} \\
            \bottomrule
        \end{tabular}
    }
    \captionof{table}{Comparison between previous AES benchmarks and our proposed \dataset. The cells highlighted in \colorbox{red!20}{red} indicate the highest number for \textit{\#Topics} and \textit{\#Traits} columns, and the unique modality for \textit{Modality} column.}
    \label{tab:AES datasets}
    \vspace{-2mm}
\end{minipage}
\end{figure}

\subsection{AES Datasets}
Existing AES datasets have advanced the field but remain some limitations (shown in Table \ref{tab:AES datasets}) \cite{Ke2019survey,li2024recent,li2024reflection}. For example, $\text{ASAP}_{\text{AES}}$ is notable for its size, enabling high-performance prompt-specific systems \cite{cozma2018ASAP}. However, differing score ranges across prompts and heavy preprocessing (\textit{e.g.}, removal of paragraph structures and named entities) reduce its utility. ASAP++ is an extension of ASAP that introduces trait-specific scores \cite{mathias2018asap++,li2024reflection}. However, its traits are coarse-grained, with all content-based traits (\textit{e.g}., coherence, persuasiveness, and thesis clarity) grouped into a single "CONTENT" category. The CLC-FCE dataset includes holistic scores and linguistic error annotations, supporting grammatical error detection alongside scoring tasks, but the small number of essays per prompt hinders the development of prompt-specific systems \cite{yannakoudakis2011CLCFCE,li2024recent}. TOEFL11 dataset focuses on native language identification and provides only coarse-grained proficiency labels (low, medium \& high), which do not fully capture essay quality. ICLE \cite{granger2009icle} and ICLE++ \cite{li2024icle++} datasets provide some of the most detailed trait-specific annotations, with ICLE++ scoring essays on 10 dimensions of writing quality. Nevertheless, these datasets are still constrained by limited topic diversity. Similarly, The AAE corpus includes 102 persuasive essays and only focuses on argument structure \cite{stab2014AAE}. To address the aforementioned limitations, we propose the \dataset benchmark, which features multimodal context, 125 unique essay topics, and comprehensive scoring across 10 distinct traits.

\subsection{AES Systems}
AES research focuses on three main categories: heuristic approaches, machine learning approaches, and deep learning approaches \cite{li2024reflection}. Heuristic AES approaches focus on holistic scoring by combining trait scores such as Organization, Coherence, and Grammar into a weighted sum. Trait-specific scores are computed using rules, like assessing Organization based on a five-paragraph format \cite{Attali2006erator}. Machine learning approaches (\textit{e.g.}, Logistic Regression and Support Vector Machine) rely on handcrafted features, such as lexical \cite{chen2013lexical}, length-based \cite{Sowmya2016linguistic,yannakoudakis2012lenthbased}, and discourse features \cite{yannakoudakis2012lenthbased}, and perform well in within-prompt scoring but struggle with generalization to new prompts. Deep learning approaches, particularly those using Transformer architectures like BERT \cite{wang2022bert}, have advanced AES by learning essay representations directly from text, enabling multi-trait and cross-prompt scoring. Among these, LLM-based approaches stand out for their ability to leverage commonsense knowledge and understand complex instructions \cite{MIZUMOTO2023llmbased}. By using prompts, LLMs can perform AES in zero-shot settings with rubrics alone \cite{lee2024TOEFL11} or in few-shot settings with minimal labeled data \cite{mansour2024fewshoting,xiao2024fewshoting}. These methods enhance flexibility, scalability, and performance, especially in low-resource scenarios.

\subsection{Multimodal Large Language Models}
MLLMs have brought significant advancements to diverse tasks and applications \cite{xi2023rise,huo2024mmneuron,yan2024urbanclip,yan2024georeasoner,zou2025deep,dang2024explainable}. Proprietary MLLMs such as GPT-4o \cite{openai2024gpt4ocard} and Gemini-1.5 \cite{gemini} have shown remarkable capabilities in multimodal challenges, excelling in areas such as multimodal reasoning and QA \cite{chang2024survey,yan2024errorradar,yan2024survey,zheng2024reefknot,yan2025position}. At the same time, Open-source MLLMs have made considerable strides. For instance, LLaVA-NEXT \cite{liu2024llavanext} utilizes a pretrained vision encoder to generate visual embeddings, which are then aligned with text embeddings through a lightweight adapter, enabling effective multimodal understanding. Similarly, MLLMs such as Qwen2-VL \cite{qwen2vl}, DeepSeek-VL \cite{lu2024deepseek}, InternVL \cite{internvl2,internvl2.5}, MiniCPM \cite{hu2024minicpmunveilingpotentialsmall}, Ovis \cite{lu2024ovisstructuralembeddingalignment}, LLaMA3 \cite{grattafiori2024llama3herdmodels} and Yi-VL \cite{young2024yi} implement innovative projection techniques to combine visual and textual features effectively, enabling many multimodal applications. These models showcase the growing potential of MLLMs in advancing both research and practical applications that rely on multimodal data \cite{qu2025tool,zou2024look,zhou2024mitigating,huang2024miner}. Therefore, we introduce \dataset, a novel benchmark designed to evaluate MLLMs’ capability to score essays with multimodal context, paving the way for AGI systems \cite{xiao2024automation,tate2024can,yan2024practical,yan2025mathagent}.
\vspace{-2mm}
\section{Dataset}
\begin{figure*}[!t]
    \centering
    \includegraphics[width=\linewidth,scale=1.00]{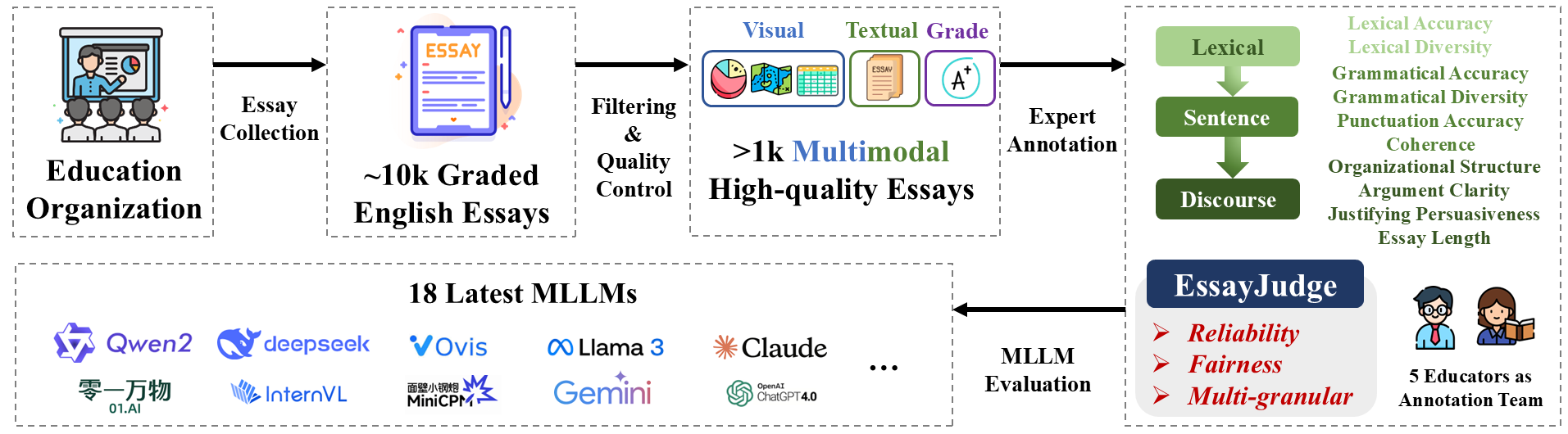}
    \caption{Roadmap illustration of \dataset dataset collection, construction and annotation.}
    \vspace{-5mm}
    \label{fig:essayjudge_dataset_roadmap}
\end{figure*}
\vspace{-2mm}
\subsection{Data Collection}
 This section describes the process of constructing our dataset to ensure high-quality data for analysis, as illustrated in Figure \ref{fig:essayjudge_dataset_roadmap}. Unlike traditional datasets that often rely on publicly available sources or textbook modifications, the data for this study originates from a K-12 Education Organization. This organization provides a repository of essays graded by experienced educators, guaranteeing the credibility and reliability.

From the original dataset, four primary fields were retained: (i) \textit{Image}, which contains the image of the writing Topics; (ii) \textit{Question}, which contains the text of the writing prompt; (iii) \textit{Essay}, representing the student’s written work; (iv) \textit{Overall Score}, reflecting the final assessment provided by professional educators. To enhance the dataset’s quality, a series of processing and cleaning steps were applied. These steps included removing essays with incomplete or low-quality responses and selecting topics that met the criteria for reliability and diversity. Through this rigorous process, we curated the \dataset dataset consisting of 1,054 multimodal essays with 125 topics. These essays, covering a broad spectrum of writing abilities, form a solid foundation for facilitating AES research.
\vspace{-2mm}
\subsection{Data Annotation Scheme}
Through discussion with English teachers and linguistic experts, we identified ten traits of essays and categorized them into three levels of granularity, progressing from fine-grained lexical features to broader sentence-level structures and, finally, to discourse-level characteristics. This hierarchical structure reflects a natural flow from the smallest units of language to the overall coherence and persuasiveness of an essay, providing a comprehensive framework for evaluation. The rubrics (with a scale of 0 to 5) can be seen in Appendix \ref{app:rubrics}. Higher scores indicate a stronger performance.

At the \textbf{lexical level}, the focus is on the precision and diversity of word usage, which forms the foundation of effective expression. Traits including \ding{182} \textit{lexical accuracy} and \ding{183} \textit{lexical diversity} assess how well the writer uses vocabulary to convey meaning, including the correctness of word choice, spelling, and semantic appropriateness, as well as the variety and richness of vocabulary demonstrated in the essay. These fine-grained features ensure that the language is both accurate and expressive.

Moving to the \textbf{sentence level}, the evaluation shifts to the internal quality of sentences and the connections between them. Traits including \ding{184} \textit{grammatical accuracy} and \ding{185} \textit{grammatical diversity} examine the correctness and variety of grammatical structures, reflecting the writer’s ability to construct well-formed and diverse sentences. \ding{186} \textit{Punctuation accuracy} ensures that punctuation marks are used appropriately to enhance clarity and readability. Additionally, \ding{187} \textit{coherence} is assessed at this level, focusing on how smoothly sentences connect through effective transitions, logical relationships, and appropriate use of conjunctions. This intermediate granularity highlights the writer’s ability to build logical and linguistically sound sentence structures that support the flow of ideas.

At the \textbf{discourse level}, the evaluation considers the overall structure, argumentation, and coherence of the essay as a whole. Traits including \ding{188} \textit{organizational structure} assess how well the essay is organized across its introduction, body, and conclusion, ensuring ideas are logically and clearly presented. \ding{189} \textit{Argument clarity} evaluates the explicitness and focus of the central argument, while \ding{190} \textit{justifying persuasiveness} measures the strength of evidence and reasoning provided to support the argument. Finally, \ding{191} \textit{essay length} ensures that the essay meets the required length while maintaining depth and focus. This broader granularity captures the writer’s ability to integrate all elements into a cohesive and persuasive whole.

\vspace{-2mm}
\subsection{Datasets Annotation Procedure}
To ensure a thorough and objective evaluation of the traits, we enlisted two experienced experts in English education, who independently assessed all 10 traits for each essay. After scoring, we compared the results and calculated the differences between the two sets of scores. For traits where the score difference was less than or equal to 1, we took the average of the two scores to establish the ground-truth score. In cases where the score difference exceeded 1, we asked another independent team consisting of three senior annotators to review the essays and discuss the traits with the team. They finally reached a consensus on the final ground-truth score, ensuring a fair and reliable outcome.

\vspace{-2mm}
\subsection{Data Details}

\dataset dataset comprises a substantial collection of 1,054 multimodal essays designed for AES (See details in Appendix \ref{app:dataset}). The dataset is categorized based on the \textit{number of images per question}, with 66.7\% being single-image questions and the remaining 33.3\% multi-image questions. 



\clearpage
\section{Experiment and Analysis}
\subsection{Experimental Setup}
\paragraph{Evaluation Groups.}
We meticulously categorized diverse MLLMs into distinct groups to assess their capabilities in trait-specific AES. (i) The \textbf{Open-Source MLLMs} category encompassed models such as Yi-VL \cite{young2024yi}, Qwen2-VL \cite{qwen2vl}, DeepSeek-VL \cite{lu2024deepseek}, LLaVA-NEXT \cite{liu2024llavanext}, InternVL2 \cite{internvl2}, InternVL2.5 \cite{internvl2.5}, MiniCPM-V2.6 \cite{hu2024minicpmunveilingpotentialsmall}, MiniCPM-LLaMA3-V2.5 \cite{hu2024minicpmunveilingpotentialsmall}, Ovis1.6-Gemma2 \cite{lu2024ovisstructuralembeddingalignment}, and LLaMA-3.2-Vision \cite{grattafiori2024llama3herdmodels}, each demonstrating their unique strengths and capabilities in multi-granular essay scoring. (ii) The \textbf{Closed-Source MLLMs} featured proprietary models like Qwen-Max \cite{qwenmax}, Step-1V \cite{step1v}, Gemini-1.5-Pro \cite{gemini}, Gemini-1.5-Flash \cite{geminiflash}, Claude-3.5-Haiku \cite{claude35h}, Claude-3.5-Sonnet \cite{claude35s}, GPT-4o-mini \cite{openai2024gpt4omini}, and GPT-4o \cite{openai2024gpt4ocard}, providing a comparison point for the performance of models that are not publicly accessible. (iii) Lastly, the \textbf{Human Performance} category served as a benchmark for human-level intelligence, enabling us to assess how closely MLLMs emulate human cognitive abilities  (More details in Appendix \ref{app:human}). The detailed prompts for MLLMs and sources of MLLMs are provided in Appendix \ref{app:prompt} and \ref{app:sources}.

\paragraph{Evaluation Metric.}
We employ Quadratic Weighted Kappa (QWK) \cite{Ke2019survey,li2024recent,li2024reflection} as our metric for scoring the similarity, which is widely used to evaluate the agreement between model scores and the ground truth. Its formula is expressed as:
\[
k = 1 - \frac{\sum_{i,j}w_{i,j}O_{i,j}}{\sum_{i,j}w_{i,j}E_{i,j}},
\]
where $w_{i,j}=\frac{(i - j)^2}{\left(N - 1\right)^2}$ is the weight matrix penalizing larger differences between $i$ and $j$, $\boldsymbol{O}_{i,j}$ is the observed agreement, and $\boldsymbol{E}_{i,j}$ is the expected agreement under random chance. QWK values range from -1 (complete disagreement) to 1 (perfect agreement). Higher values are expected. 

\subsection{Main Results}

\begin{figure}[htbp]
  \centering
  \includegraphics[width=0.5\textwidth]{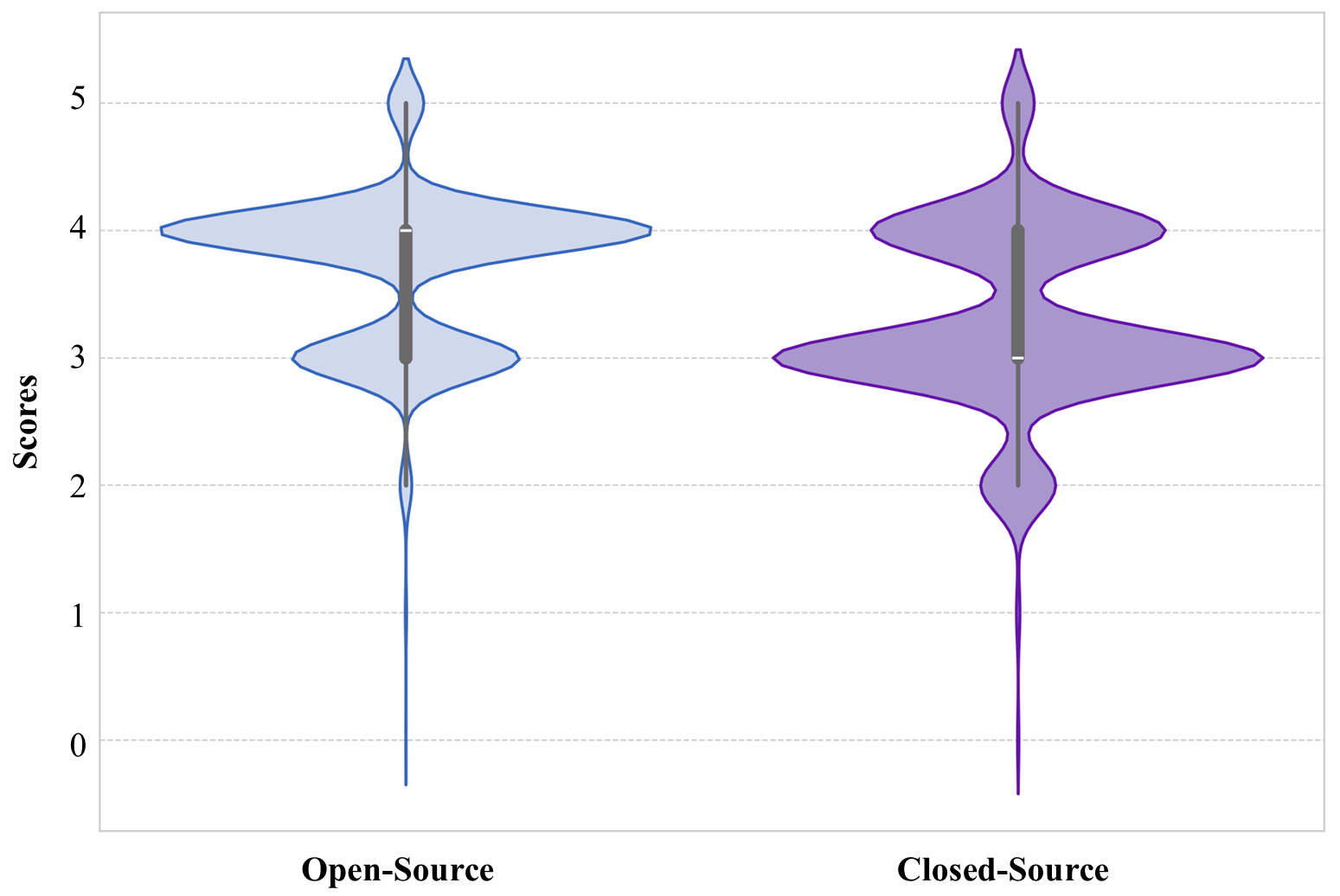}
  \caption{The open-source and closed-source MLLMs' distribution of average scores among ten traits.}
  \label{fig:violin}
  \vspace{-4mm}
\end{figure}

\textbf{Closed-source MLLMs demonstrate significant superiority over open-source MLLMs in essay scoring tasks, with GPT-4o achieving the strongest overall performance.} Table \ref{tab:main result} illustrates that closed-source MLLMs consistently outperform open-source MLLMs across ten traits. This advantage is likely due to the high-quality proprietary datasets and advanced training techniques leveraged by closed-source models \cite{yu2024large_privary_data,wang2023large_privary_data}, enabling them to achieve more balanced and robust performance. GPT-4o stands out as the best-performing model, achieving the highest QWK scores across multiple traits except for Argument Clarity, highlighting its exceptional capability. Among open-source MLLMs, InternVL2 emerges as the best performer. However, its overall performance remains behind closed-source ones, underscoring the gap between open-source and closed-source models in terms of capturing complex and evaluative subtleties.

\begin{table*}[htbp]
\vspace{-3mm}
\centering
\small
\renewcommand\tabcolsep{3pt} 
\renewcommand\arraystretch{1.1} 
\resizebox{1.0\linewidth}{!}{
    \begin{tabular}{
        @{} 
        c | c 
        | >{\centering\arraybackslash}m{0.8cm} 
        >{\centering\arraybackslash}m{0.8cm} 
        | >{\centering\arraybackslash}m{0.8cm} 
        >{\centering\arraybackslash}m{0.8cm} 
        >{\centering\arraybackslash}m{0.8cm} 
        >{\centering\arraybackslash}m{0.8cm} 
        | >{\centering\arraybackslash}m{0.8cm} 
        >{\centering\arraybackslash}m{0.8cm} 
        >{\centering\arraybackslash}m{0.8cm} 
        >{\centering\arraybackslash}m{0.8cm} 
        @{} 
    }
    \toprule
    \multirow{2}{*}{\textbf{MLLMs}} & \multirow{2}{*}{\textbf{\#Para.}} &
    \multicolumn{2}{c|}{\textbf{Lexical Level}} &
    \multicolumn{4}{c|}{\textbf{Sentence Level}} &
    \multicolumn{4}{c}{\textbf{Discourse Level}} \\ 
    \cmidrule(lr){3-4} \cmidrule(lr){5-8} \cmidrule(lr){9-12}
     &  & \textbf{LA} & \textbf{LD} 
     & \textbf{CH} & \textbf{GA} & \textbf{GD} & \textbf{PA} 
     & \textbf{AC} & \textbf{JP} & \textbf{OS} & \textbf{EL} \\
    \midrule
    \multicolumn{12}{c}{\textit{Open-Source MLLMs}} \\
    \midrule
    Yi-VL \cite{young2024yi} & 6B & 0.07 & 0.05 & 0.08 & 0.09 & 0.05 & 0.09 & 0.05 & 0.13 & 0.05 & 0.07\\
    Qwen2-VL \cite{qwen2vl} & 7B & 0.20 & 0.26 & 0.13 & 0.21 & 0.16 & 0.12 & 0.17 & 0.10 & 0.14 & 0.15  \\
    DeepSeek-VL \cite{lu2024deepseek} & 7B & 0.09 & 0.12 & 0.12 & 0.13 & 0.35 & 0.06 & 0.18 & 0.21 & 0.08 & 0.09\\
    LLaVA-NEXT \cite{liu2024llavanext} & 8B & 0.02 & 0.04 & 0.03 & 0.11 & 0.10 & 0.02 & 0.12 & 0.15 & 0.02 & 0.10\\
    InternVL2 \cite{internvl2} & 8B & 0.28 & 0.27 & 0.34 & 0.36 & 0.31 & 0.33 & 0.25 & 0.29 & 0.31 & 0.29 \\
    InternVL2.5 \cite{internvl2.5} & 8B & 0.14 & 0.29 & 0.29  & 0.29 & 0.31 & 0.26 & 0.15 & 0.21 & 0.25 & 0.22\\ 
    MiniCPM-V2.6 \cite{hu2024minicpmunveilingpotentialsmall} & 8B & 0.18 & 0.07 & 0.08 & 0.16 & 0.09 & 0.04 & 0.12 & 0.35 & 0.06 & 0.24 \\
    MiniCPM-LLaMA3-V2.5 \cite{hu2024minicpmunveilingpotentialsmall} & 8B & 0.37 & 0.27 & 0.36 & 0.29 & 0.34 & 0.29 & 0.09 & 0.18 & 0.21 & 0.09\\
    Ovis1.6-Gemma2 \cite{lu2024ovisstructuralembeddingalignment} & 9B & 0.15 & 0.11 & 0.13 & 0.39 & 0.27 & 0.36 & 0.11 & 0.13 & 0.14 & 0.21\\ 
    LLaMA-3.2-Vision \cite{grattafiori2024llama3herdmodels} & 11B & 0.20 & 0.16 & 0.17 & 0.14 & 0.11 & 0.12 & 0.09 & 0.17 & 0.17 & 0.16\\ 
    \midrule
    \multicolumn{12}{c}{\textit{Closed-Source MLLMs}} \\
    \midrule
    Qwen-Max \cite{qwenmax} & - & 0.57 & 0.51 & 0.52 & 0.56 & 0.48 & 0.40 & 0.34 & 0.54 & 0.45 & 0.41 \\ 
    Step-1V \cite{step1v} & - & 0.52 & 0.40 & 0.49 & 0.50 & 0.46 & 0.37 & 0.26 & 0.39 & 0.31 & 0.25\\
    Gemini-1.5-Pro \cite{gemini} & - & 0.52 & 0.46 & 0.57 & 0.56 & 0.51 & 0.35 & 0.29 & 0.46 & \high{0.54} & 0.28\\
    Gemini-1.5-Flash \cite{geminiflash} & - & 0.46 & 0.40 & 0.48 & 0.53 & 0.41 & 0.33 & 0.33 & 0.42 & 0.47 & 0.28  \\ 
    Claude-3.5-Haiku \cite{claude35h} & - & 0.59 & 0.54 & 0.53 & 0.50 & 0.57 & 0.40 & \best{0.35} & 0.39 & 0.48 & 0.33\\
    Claude-3.5-Sonnet \cite{claude35s} & - & \high{0.66} & \high{0.60} & \high{0.58} & \high{0.66} & \high{0.60} & \high{0.57} & 0.33 & 0.46 & 0.39 & 0.35 \\
    GPT-4o-mini \cite{openai2024gpt4omini} & - & 0.64 & 0.56 & 0.54 & 0.58 & 0.54 & 0.45 & \high{0.33} & \high{0.57} & 0.45 & \high{0.46}\\
    GPT-4o \cite{openai2024gpt4ocard} & - & \best{0.89} & \best{0.89} & \best{0.87} & \best{0.85} & \best{0.61} & \best{0.65} & 0.30 & \best{0.80} & \best{0.79} & \best{0.70} \\
    \midrule
    \multicolumn{12}{c}{\textit{Human Performance}} \\
    \midrule
    Human performance & - & 0.91 & 0.91 & 0.89 & 0.93 & 0.56 & 0.86 &0.72 & 0.86 & 0.88 & 0.77 \\   
    \bottomrule
    \end{tabular}
}
\caption{Comparison of open-source and closed-source MLLM performance (QWK). The highest and second highest scores among MLLMs in each column are highlighted in \colorbox{red!20}{red} and \colorbox{blue!20}{blue}, respectively.}
\vspace{-3mm}
\label{tab:main result}
\end{table*}

\begin{figure*}[htbp]
  \centering
  \includegraphics[width=1\textwidth]{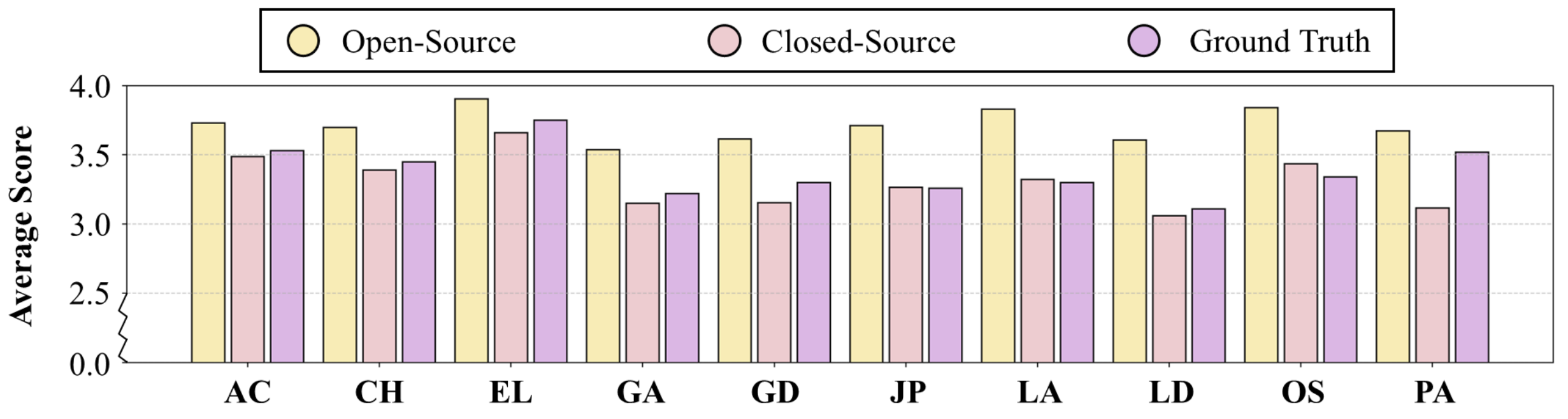}
  \caption{The average scores of open-source MLLMs, closed-source MLLMs and ground-truth across ten traits.}
\label{fig:average_scores}
\end{figure*}

\textbf{Closed-source MLLMs exhibit distinct scoring patterns compared to open-source counterparts, characterized by greater score variability and stricter adherence to rubrics.} As revealed in Figure \ref{fig:violin}, closed-source models demonstrate significantly higher score variance (0.49 vs. 0.34), suggesting enhanced capacity to differentiate essay quality through broader score distribution across performance levels. This contrasts with open-source models' tendency to cluster scores in the mid-range (3-4 points), reflecting limited discriminative capacity for quality extremes. As shown in Figure \ref{fig:average_scores}, the scoring rigor of closed-source systems is further evidenced by their consistent assignment of lower scores across critical traits including Argument Clarity, Coherence, and Linguistic Features, with human ratings typically intermediate between the two model types. This systematic conservatism stems from closed-source models' strict adherence to quantitative rubtics \cite{kundu2024lstricter}, prioritizing error penalization (grammatical inaccuracies, logical inconsistencies) and standardization over nuanced language appreciation. While ensuring transparency and reproducibility, this approach may undervalue creative language use, contrasting with human raters' contextual flexibility and open-source models' reduced capacity to assess linguistic complexity effectively.

\begin{figure}[htbp]
  \centering
  \includegraphics[width=0.5\textwidth]{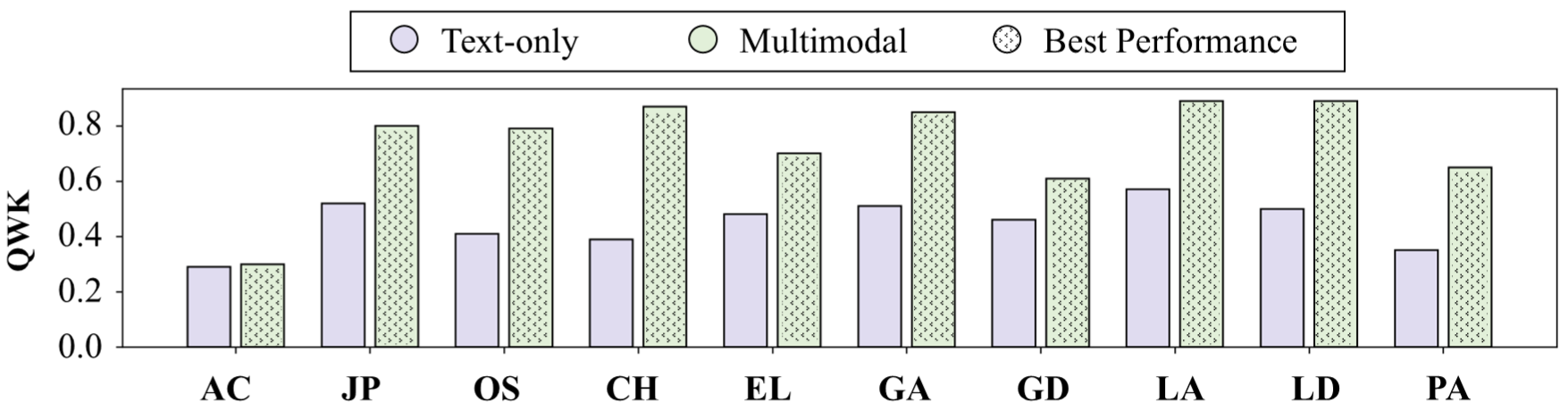}
  \caption{GPT-4o’s QWK values across traits for text-only and multimodal settings.}
\label{fig:gpt4o_without_image}
\vspace{-2mm}
\end{figure}

\textbf{Multimodal inputs play a critical role in improving the performance of LLMs in AES tasks.} Our ablation study compared the performance of models in two settings, including multimodal inputs (text and images) versus text-only inputs. As shown in Figure \ref{fig:gpt4o_without_image}, when image information was removed, GPT-4o experienced a decrease in QWK scores across all ten traits. These findings underscore that visual features provide essential evaluative dimensions that are inaccessible to text-only approaches, especially when images contain critical information supporting the argument. Additional evaluation results for other models are provided in the Appendix \ref{ablation}.

\vspace{-2mm}
\subsection{Trait-Specific Analysis}

\begin{figure}[!t]
  \centering
  \includegraphics[width=0.5\textwidth]{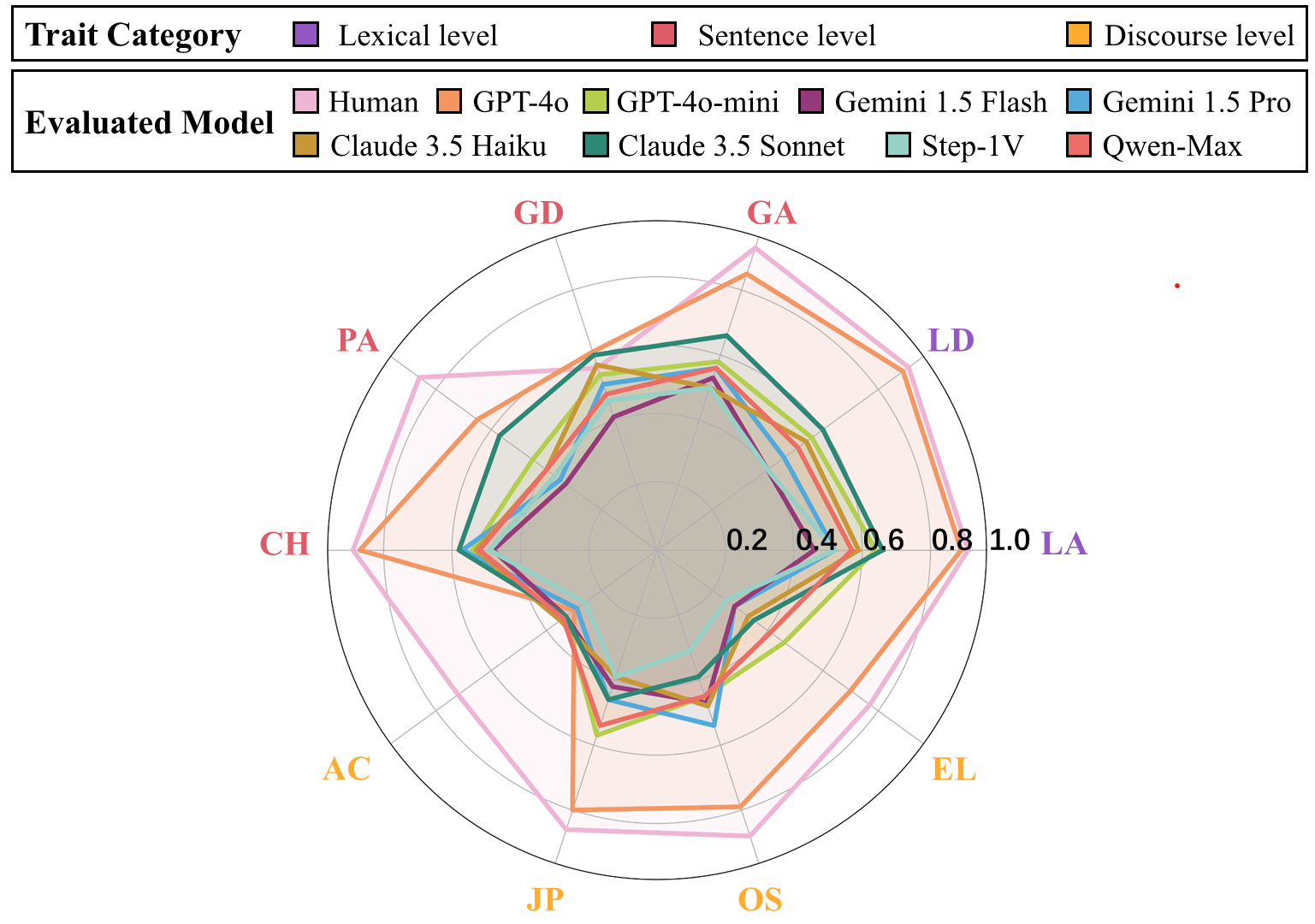}
  \caption{The average QWK values of representative models across three granularities of ten traits.}
  \label{fig:qwk_closed}
  \vspace{-2mm}
\end{figure}

\textbf{Closed-source MLLMs perform poorly in evaluating Argument Clarity and Essay Length while excelling at lexical-level assessment.} As shown in Figure \ref{fig:qwk_closed}, their low QWK values for Argument Clarity and Essay Length are because closed-source LLMs rely heavily on surface-level features like grammar and vocabulary, making them ineffective in handling complex arguments that require contextual understanding and reasoning. For Essay Length, they often hallucinate word counts \cite{rawte2023Hallucination} and over-rely on quantitative measures, leading to an overvaluation of verbosity and an undervaluation of concise but effective writing \cite{jeon2021countering}. In contrast, closed-source MLLMs' strong lexical performance is due to exposure to a extensive, high-quality datasets \cite{shi2023detecting_privary_data}, enabling superior precision and variety in lexical choice.

\begin{figure}[htbp]
  \centering
  \includegraphics[width=0.5\textwidth]{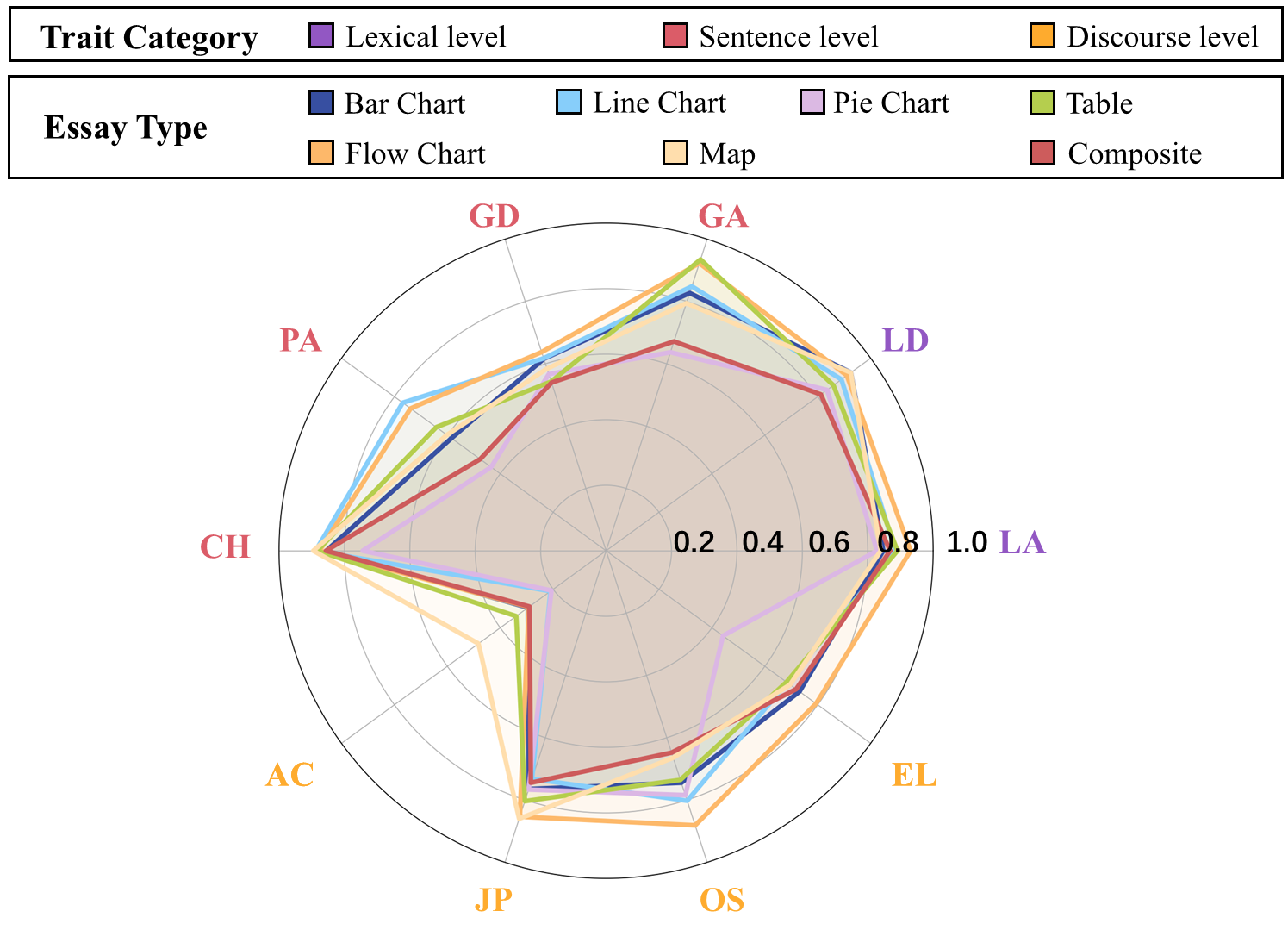}
  \caption{GPT-4o's mean QWK values across different essay types across three granularities of ten traits.}
  \label{fig:radorgpt4o}
  \vspace{-2mm}
\end{figure}

\textbf{MLLMs demonstrate outstanding performance in assessing Coherence when grading essays related to line charts.} For example, as shown in Figure \ref{fig:radorgpt4o}, GPT-4o achieves a high QWK value of 0.89 for the coherence trait when evaluating line chart essays. The results of the top three open-source and top three closed-source MLLMs are provided in the appendix \ref{app:essay type}. This is primarily because line charts have a highly linear information structure that emphasizes trends and changes, making the logical relationships between data points clear and easy to follow \cite{islam2024chart}. Additionally, the core information in line chart descriptions typically focuses on key points such as turning points, peaks, and troughs, which simplifies the logical chain and allows models to effectively capture and evaluate the coherence of the text.

\begin{figure}[!t]
  \centering
  \includegraphics[width=0.5\textwidth]{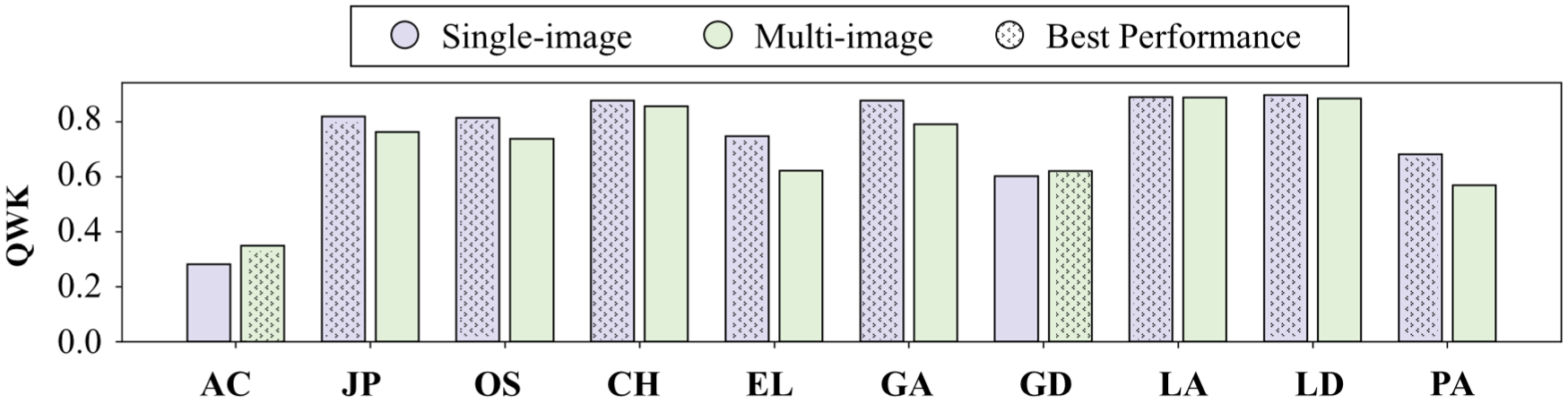}
  \caption{Comparison of GPT-4o's QWK values across ten traits between single-image and multi-image settings.}
  \label{fig:gpt4o_single_multi}
  \vspace{-2mm}
\end{figure}

\begin{figure}[!t]
  \centering
  \includegraphics[width=0.5\textwidth]{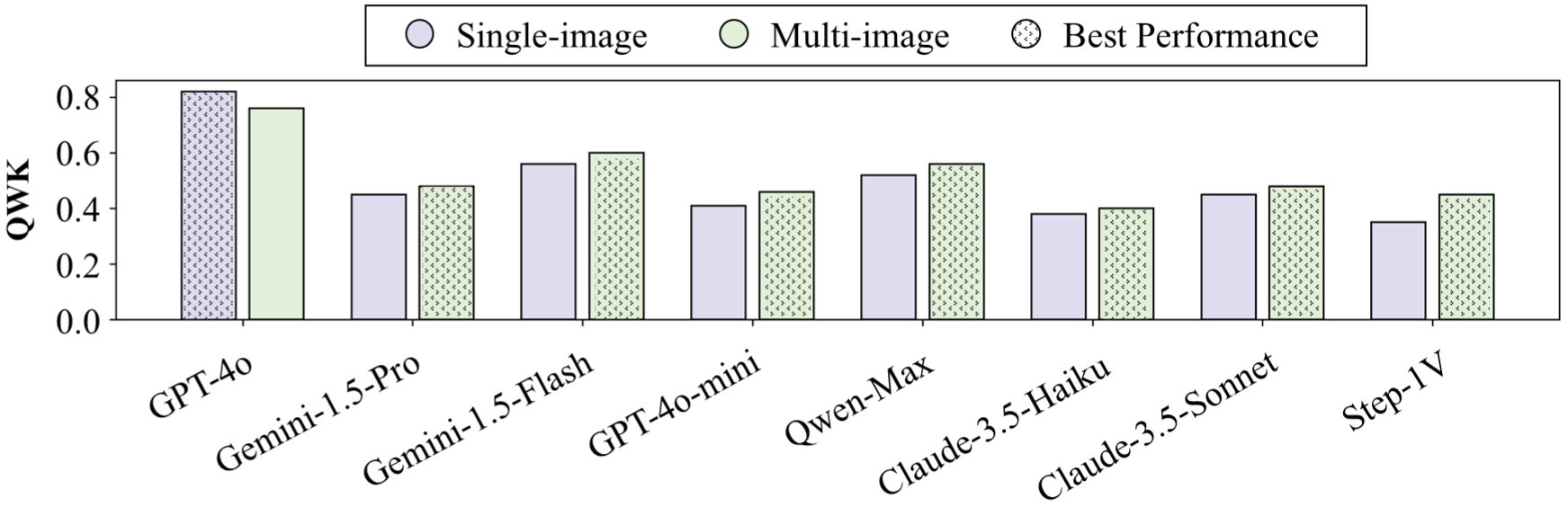}
  \caption{The \textbf{closed-source} MLLMs' QWK values of JP trait among single-image \& multi-image settings.}
  \label{fig:bar_chart_JP_closed}
  \vspace{-2mm}
\end{figure}

\begin{figure}[!t]
  \centering
  \includegraphics[width=0.5\textwidth]{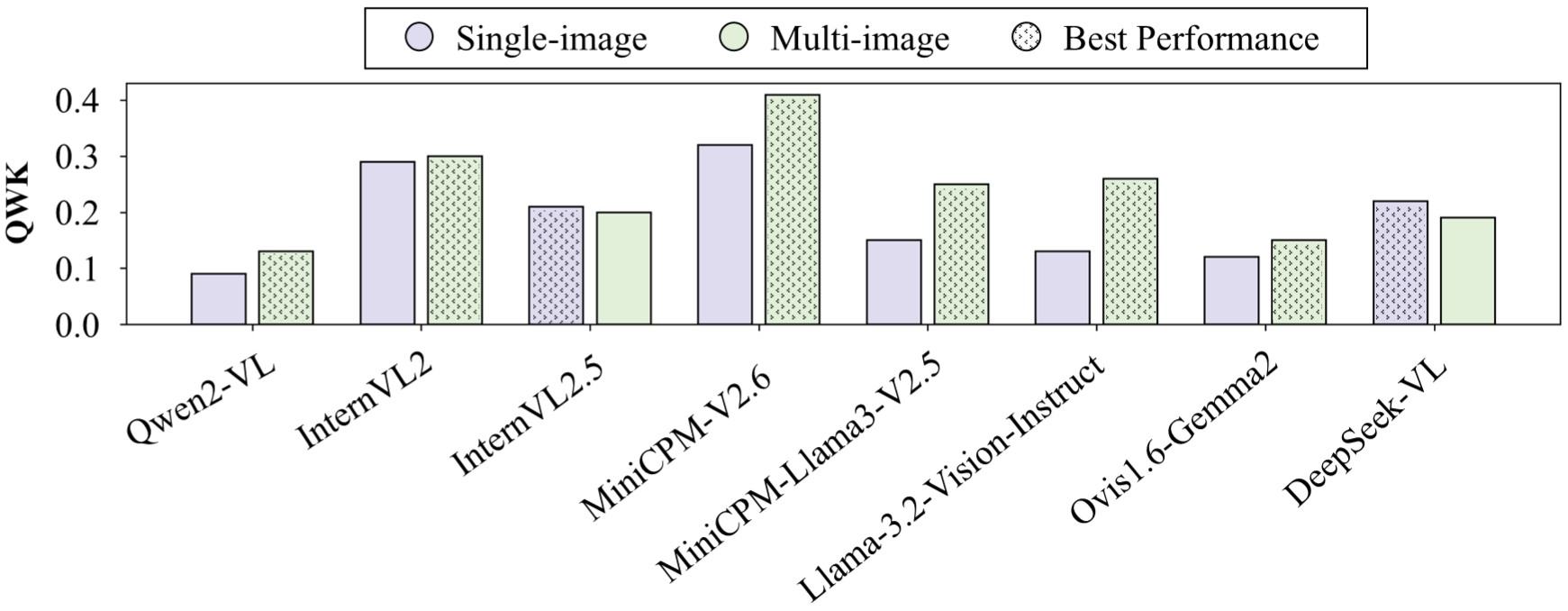}
  \caption{The \textbf{open-source} MLLMs' QWK values of JP trait among single-image \& multi-image settings.}
  \label{fig:bar_chart_JP_open}
  \vspace{-4mm}
\end{figure}

\begin{figure*}[!t]
  \centering
  \includegraphics[width=\textwidth]{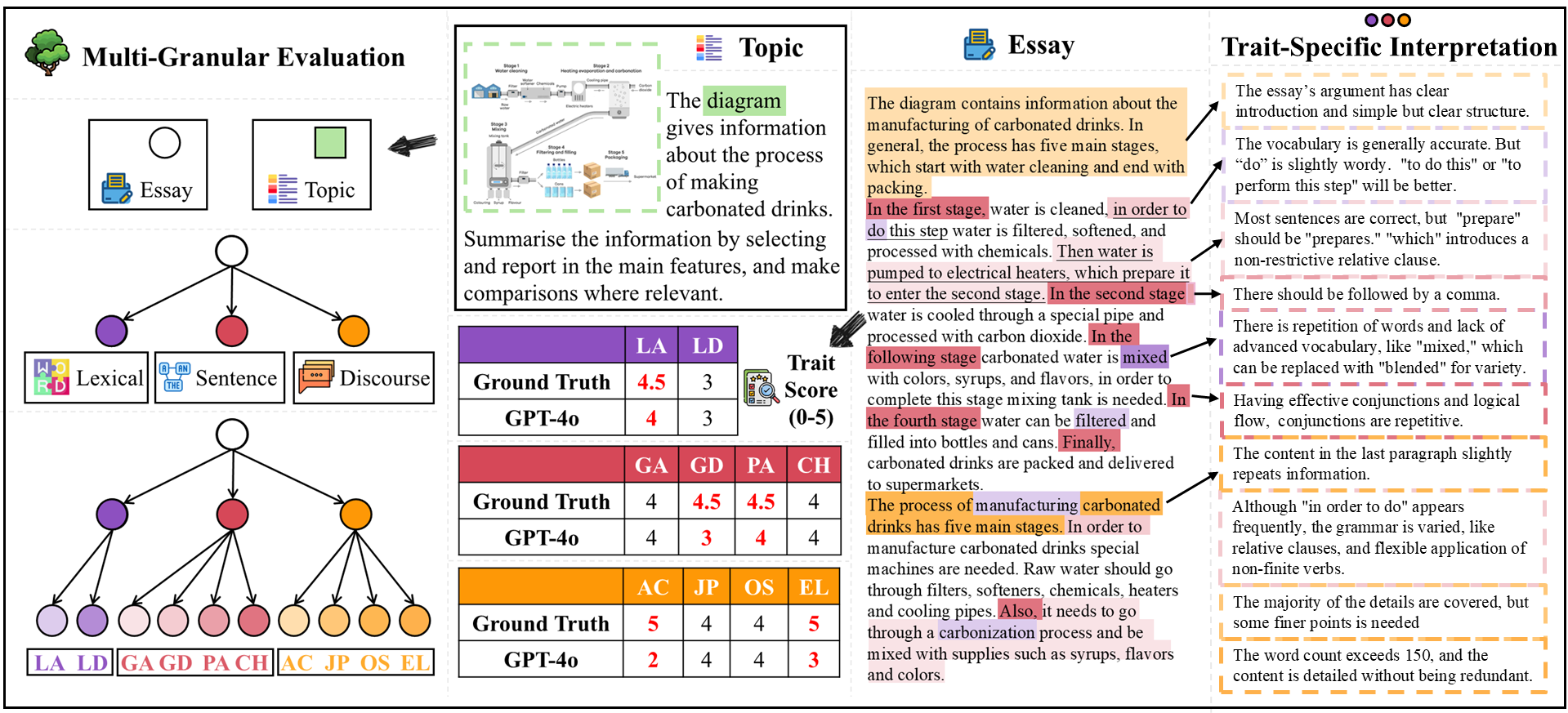}
  \caption{The example case of multi-granular evaluation for the essay.}
\label{fig:case_study}
\end{figure*}

\subsection{Analysis of \#image Setting}
\textbf{Most closed-source MLLMs perform better in evaluating essays with single-image setting.} As shown in Figure \ref{fig:gpt4o_single_multi}, GPT-4o performed better on single-image tasks except for JP and GD traits. Appendix \ref{app:single_multi} shows that among all evaluated closed-source MLLMs, only three models do not follow this pattern. Single-image tasks are simpler and more focused, typically requiring students to describe one logical theme. This clear structure makes it easier for models to capture key information and provide accurate evaluations, without the need for complex comparisons or logical integration across multiple images. In contrast, multi-image tasks involve comparing, relating, and integrating data from multiple sources, which increases task complexity and the likelihood of errors in both student responses and model evaluations. 

\textbf{For Justifying Persuasiveness, most MLLMs perform better when evaluating essays with multi-image setting.} We selected the top-performing eight models from both open-source and closed-source MLLMs, as shown in Figure \ref{fig:bar_chart_JP_closed} and Figure \ref{fig:bar_chart_JP_open}. Unlike most traits, where MLLMs tend to perform better on single-image tasks, the evaluation of Justifying Persuasiveness shows a distinct advantage in multi-image settings. This may be because multi-image tasks inherently provide richer and more diverse data points, enabling students to construct more evidence-based arguments.

\subsection{Case Study}

Figure \ref{fig:case_study} shows the detailed multi-granular evaluation for one of the essays. Additional examples are shown in Appendix \ref{app:example}. Specifically, we can find that argument clarity is the most discrepant from the ground truth. \textit{Argument clarity} is crucial in AES, as it directly reflects whether the author’s central ideas are in alignment with the essay requirement \cite{falk-lapesa-2023-bridging}. Leading models like GPT-4o show relatively poor performance in assessing argument clarity, which is illustrated in Figure \ref{tab:main result}. However, argument clarity serves as a key indicator of a model's reasoning abilities and its capacity to integrate and process complex information, including visual elements. The ability to clearly present and logically connect ideas is essential for both multimodal understanding and reasoning, making it a critical benchmark for evaluating the AES performance of MLLMs. Addressing these challenges in future MLLMs could significantly improve their ability to assess essays with complex reasoning and enhance their multimodal integration capabilities.

\section{Conclusion}

In this work, we presented \dataset, the first multimodal benchmark designed to evaluate the AES capabilities of MLLMs across lexical, sentence, and discourse-level traits. Addressing longstanding limitations in traditional AES approaches, \dataset leverages MLLMs’ inherent strengths in contextual understanding and multimodal analysis, enabling more precise, trait-specific evaluations without reliance on handcrafted features. Our comprehensive evaluation of 18 representative MLLMs highlights current limitations of MLLM-based AES systems. Notably, closed-source MLLMs such as GPT-4 demonstrate superior performance compared to open-source counterparts, yet a significant gap remains in achieving human-level accuracy. 

We envision that \dataset will not only drive innovation in AES but also serve as a stepping stone toward broader applications of MLLMs in educational assessment and beyond. The research community can address the challenges identified and foster the development of more accurate and interpretable AES systems towards AGI.

\clearpage
\section*{Limitations}
\label{sec:limitations}
Despite the findings we demonstrate in our work, there still exist minor limitations:
\begin{itemize}
\item [1.] The datasets used in this study primarily consist of essays written by non-native speakers of English, making it unclear whether our conclusions apply to essays written by native speakers, such as those in the ASAP dataset. However, since our rubrics are broadly applicable and not designed specifically for non-native speakers, we believe the conclusions can be generalized to essays written by native speakers as well.
\item [2.] Although our study covers diverse topics, including healthcare, biology, demographics, environment, education and so on, there is still a demand for a more generalized benchmark. Further expansion is needed to address a wider variety of writing contexts and disciplines, ensuring its broader applicability across different writing tasks.
\end{itemize}

\section*{Acknowledgements}
This work was supported by Open Project Program of Guangxi Key Laboratory of Digital Infrastructure (Grant Number: GXDIOP2024015); Guangdong Provincial Department of Education Project (Grant No.2024KQNCX028); Scientific Research Projects for the Higher-educational Institutions (Grant No.2024312096), Education Bureau of Guangzhou Municipality; Guangzhou-HKUST(GZ) Joint Funding Program (Grant No.2025A03J3957), Education Bureau of Guangzhou Municipality.

\bibliography{essayjudge}

\clearpage
\appendix
\section{Trait-Specific Rubrics}
\label{app:rubrics}
In this section, we introduce the rubrics used to annotate the 10 traits for each essay in our dataset. The rubrics are detailed in Table \ref{tab:AC} to Table \ref{tab:PA}. Each trait is assessed using a numerical score ranging from 0 to 5. A score of 5 represents high-quality performance with respect to the trait being evaluated, while a score of 0 represents low-quality performance in the same regard.
\section{\dataset Dataset Details}
\label{app:dataset}

\subsection{Dataset Scope}
In the \dataset benchmark, the essay requirements are all set at university-level difficulty. The essays \textit{highly depend on visual information}. Therefore, when writing the essays, students need to use the information provided by the pictures as evidence to support their arguments. This creates a unique multimodal challenge and provides a basis for evaluating the ability of MLLMs to handle diverse and complex information in the multimodal context of AES.

\subsection{Dataset Categorization}
The Figure \ref{tab:statistics} above provides a detailed breakdown of the \dataset dataset, which consists of 1054 multimodal essays. The dataset is categorized based on the type of image it includes, with 66.7\% (703 essays) containing a single image and 33.3\% (351 essays) containing multiple images. Further classification is made based on the type of visual content within these essays, with the most common type being flow charts (28.9\%), followed by bar charts (20.0\%), and tables (14.5\%). Other image types include line charts (13.8\%), pie charts (6.7\%), maps (5.9\%), and composite charts (10.2\%). This dataset provides valuable insights into the diversity of multimodal elements incorporated in essay content.

\begin{figure}[htbp]
\centering
\small
\renewcommand\tabcolsep{5pt}
\renewcommand\arraystretch{1.1} 
\resizebox{\columnwidth}{!}{ 
    \begin{tabular}{lc}
        \toprule
        \textbf{Statistic} & \textbf{Number} \\
        \midrule
        Total Multimodal Essays & 1,054 \\
        \midrule
        Image Type &  \\
        ~- Single-Image & 703 (66.7\%) \\
        ~- Multi-Image & 351 (33.3\%) \\
        \midrule
        Multimodal Essay Type &  \\
        ~- Flow Chart & 305 (28.9\%) \\
        ~- Bar Chart & 211 (20.0\%) \\
        ~- Table & 153 (14.5\%) \\
        ~- Line Chart & 145 (13.8\%) \\
        ~- Pie Chart & 71 (6.7\%) \\
        ~- Map & 62 (5.9\%) \\
        ~- Composite Chart & 107 (10.2\%) \\
        \bottomrule
    \end{tabular}
}
\captionof{table}{Key statistics of \dataset dataset.}
\label{tab:statistics}
\end{figure}

\subsection{Dataset Topic}
Our dataset includes 125 distinct essay topics, which span a wide range of themes such as population, environment, education, production, evolution, and so on. The topics represent a diverse array of subjects, offering a broad scope for analysis. In the Table \ref{tab:topic}, we highlight the top five most frequent topics in the dataset, providing an overview of the predominant themes present in the essays.

\subsection{Annotation Details}
During the annotation process, we found that when two experts independently scored for the first time, the proportion of the score difference less than or equal to 1 was 94.8\%. This data fully indicates that the scoring consistency of the two experts is very high, reflecting that the scoring of the two experts is relatively accurate. Table \ref{tab:score_difference_proportion} shows the proportions of the score difference less than or equal to 1 based on specific traits.

\begin{figure}[htbp]
\centering
\scriptsize
\renewcommand\tabcolsep{8pt}
\resizebox{\columnwidth}{!}{    
\begin{tabular}{ccc}
        \toprule
        \textbf{Traits} & \textbf{\#Essays} & \textbf{Proportion} \\
        \midrule
            AC & 900 & 85.4\% \\
            JP & 1,035 & 98.2\% \\
            OS & 1,016 & 96.4\% \\
            CH & 1,038 & 98.5\% \\
            EL & 920 & 87.3\% \\
            GA & 1,025 & 97.2\% \\
            GD & 1,044 & 99.1\% \\
            LA & 1,034 & 98.1\% \\
            LD & 1,044 & 99.1\% \\
            PA & 936 & 88.8\% \\
        \midrule
            Total & 9,992 & 94.8\%\\
        \bottomrule
    \end{tabular}
    }
\captionof{table}{The proportions of the score difference less than or equal to 1 based on specific traits.}
\label{tab:score_difference_proportion}
\end{figure}

\section{Additional Experimental Details}
\label{app:experiment}

\subsection{Analysis on Scaling Law}
\label{scaling law}
We conducted additional experiments using a larger open-source MLLM—Qwen-VL (72B). Compared to Qwen-VL (7B), this model shows a significant performance improvement, consistent with the expected scaling law (as shown in the Table \ref{tab:scaling law}). However, even with increased model size, these models still underperform compared to most closed-source MLLMs, which is in line with our conclusions.

\subsection{Multimodal Ablation Study}
\label{ablation}
\begin{figure*}[htbp]
  \centering
  \includegraphics[width=1\textwidth]{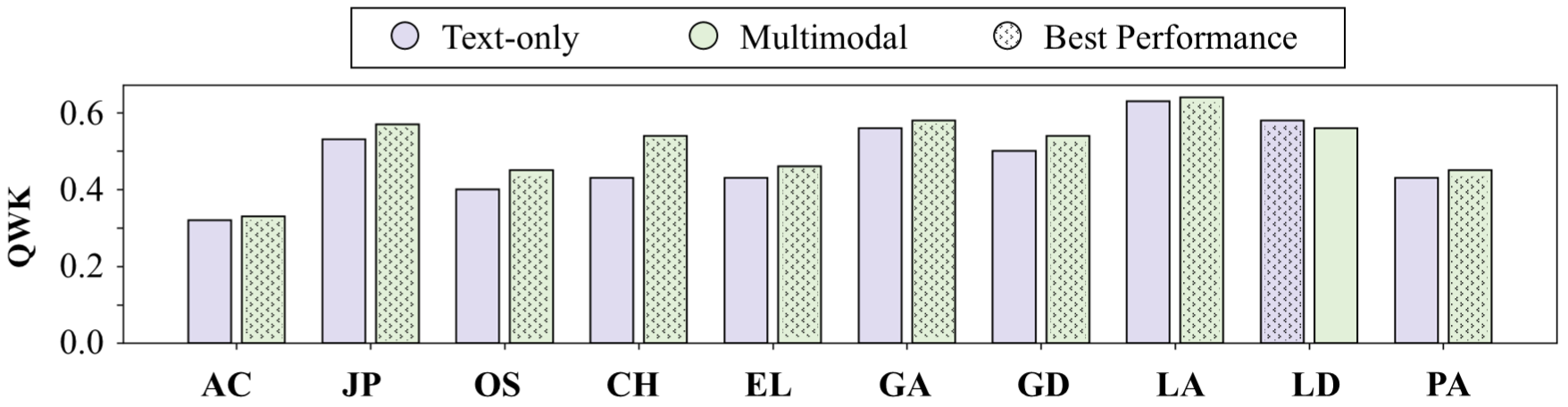}
  \caption{GPT-4o-mini’s QWK values across traits for text-only and multimodal settings.}
\label{fig:gpt4o_mini_without_image}
\end{figure*}

We also evaluated GPT-4o-mini. Figure \ref{fig:gpt4o_mini_without_image} shows that GPT-4o-mini exhibited a decline in nine out of ten traits when image information was removed, with the exception of lexical diversity. Except for GPT-4o and GPT-4o-mini, we also evaluated text-only LLM - GPT-3.5, which also showed noticeable performance drops (as shown in Table \ref{tab:ablation study}  below). This finding further underscores the importance of multimodal inputs, as visual features provide essential evaluative dimensions that text-only approaches cannot capture, especially when images contain critical supporting information.

\subsection{Human Performance Evaluation}
\label{app:human}
In the Human Performance section, four postgraduate students with excellent English backgrounds independently evaluated the essays, scoring them across ten distinct traits. To ensure the reliability of their assessments, each student was assigned an approximately equal number of the 1,054 essays, ensuring a balanced workload.

The evaluations were conducted independently, with no discussions permitted between the evaluators to preserve the integrity of the scoring process. Each postgraduate student provided their assessments based solely on their expertise and understanding of the scoring criteria.

This analysis underscores the ability of human assessments to capture the distinct traits of essays, highlighting their role in reflecting the nuances of natural intelligence. These evaluations also reveal the gap between large language models and human cognitive capabilities, serving as a benchmark for the advancements that machine intelligence strives to achieve.

\subsection{Prompt for MLLM Evaluation}
\label{app:prompt}
For the evaluation of MLLMs, we designed prompts that consist of four distinct parts: Task Definition, Rubrics, Reference Content, and Instruction. The details are shown in Figure \ref{fig:prompt example}.

The input to the model includes the question text, the accompanying image(s), the student's essay, as well as the specific trait to be evaluated and its corresponding rubrics. 

The output should be the only a numerical score, in line with the requirements set forth in the Task Definition. However, given that some models, especially the open-source MLLMs, tend to deviate from the only score task and produce outputs beyond what is expected, we explicitly reinforce the requirement for a numerical score in the final Instruction section of the prompt. This redundancy aims to ensure adherence to the evaluation task and improve the reliability of the scoring process.

\subsection{Model Sources}
\label{app:sources}
Table \ref{tab:mllm hyperparams} details specific sources for the various MLLMs we evaluated. The hyperparameters for the experiments are set to their default values unless specified otherwise.

\section{More on Trait-Specific Analysis}
\label{app:essay type}
\begin{figure*}[htbp]
  \centering
  \includegraphics[width=0.5\textwidth]{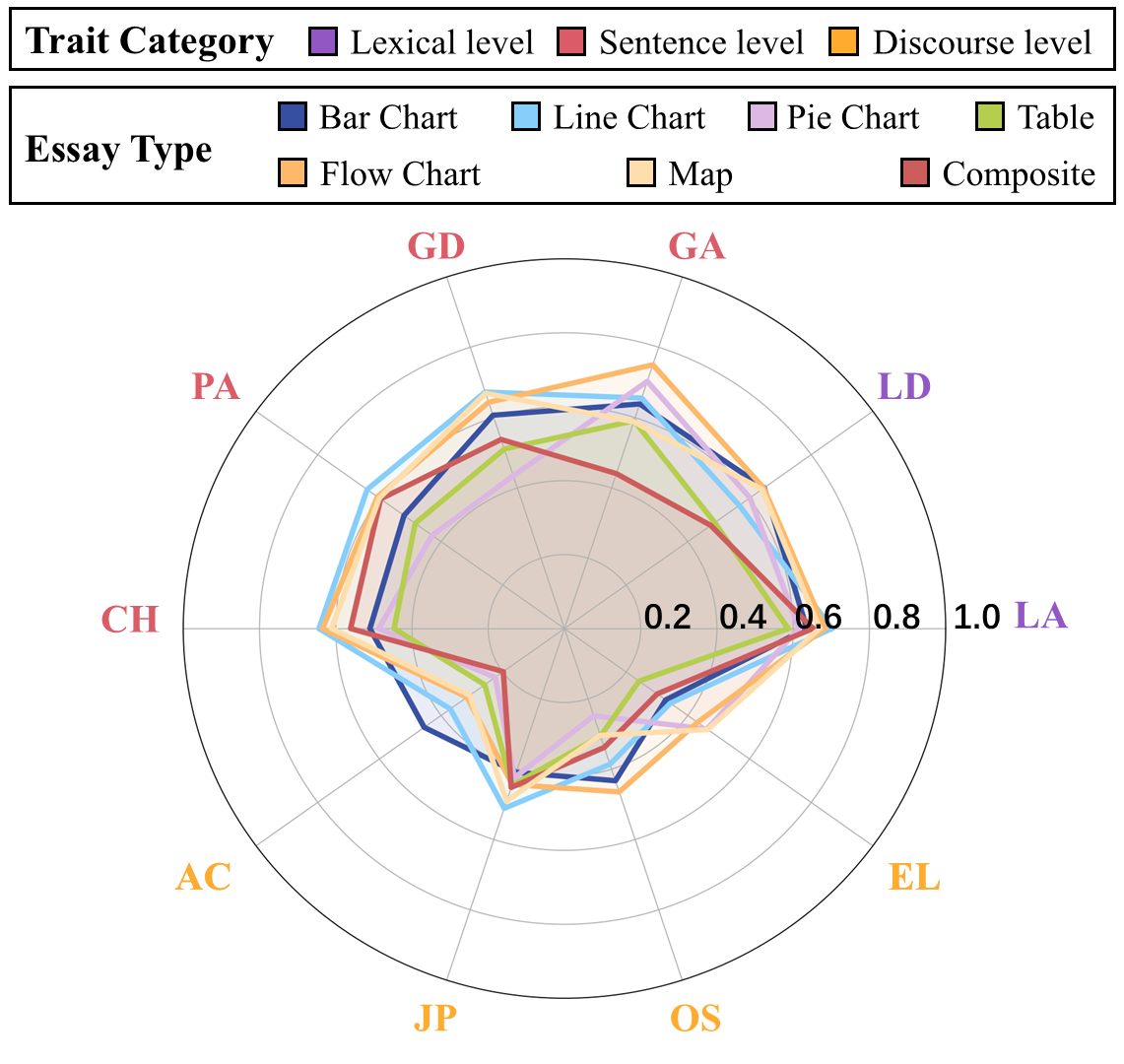}
  \caption{The relation between essay type and Claude-3.5-Sonnet's QWK.}
\label{fig:claude_sonnet_type}
\end{figure*}

\begin{figure*}[htbp]
  \centering
  \includegraphics[width=0.5\textwidth]{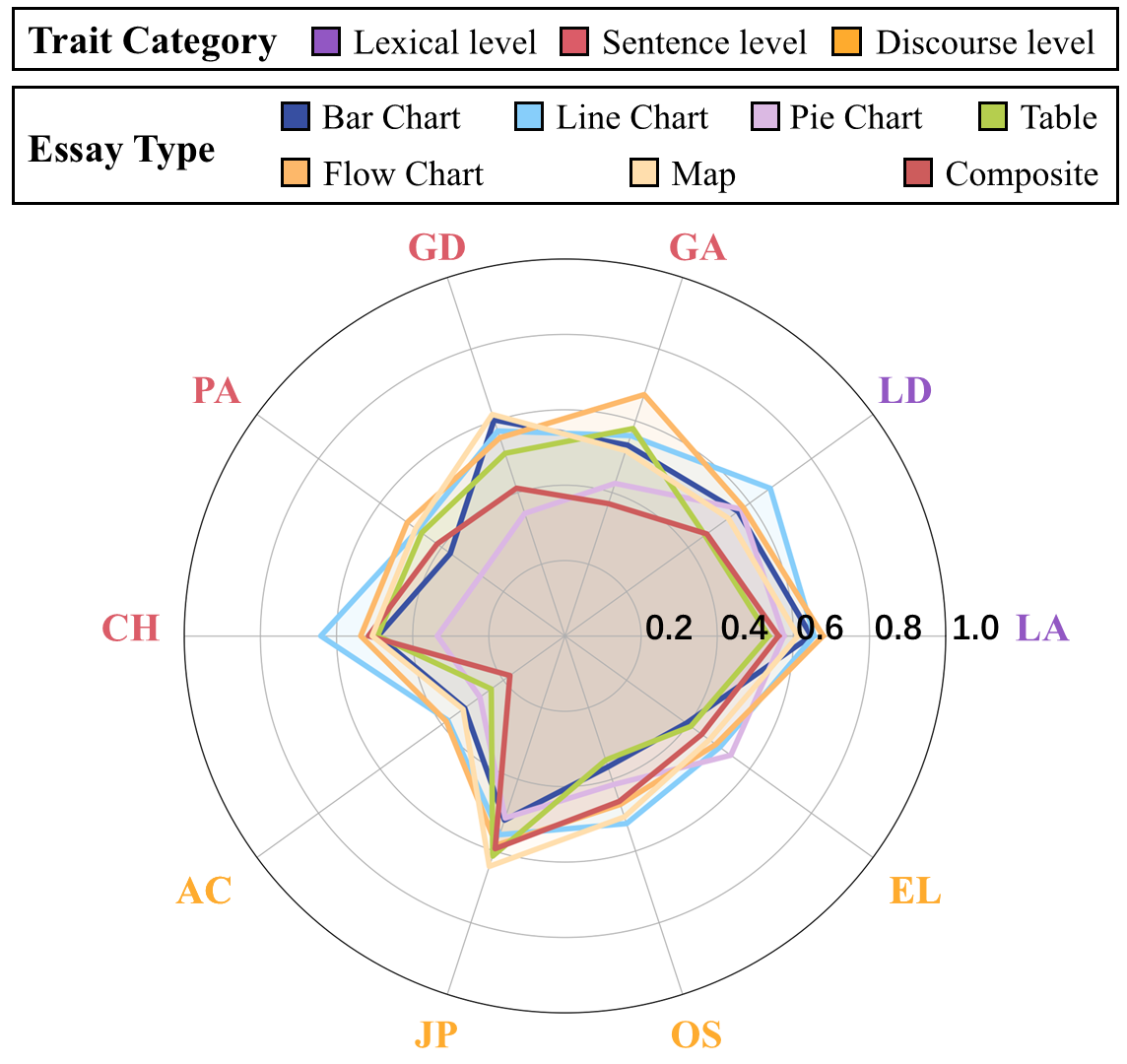}
  \caption{The relation between essay type and GPT-4o-mini's QWK.}
\label{fig:4o_mini_type}
\end{figure*}

\begin{figure*}[htbp]
  \centering
  \includegraphics[width=0.5\textwidth]{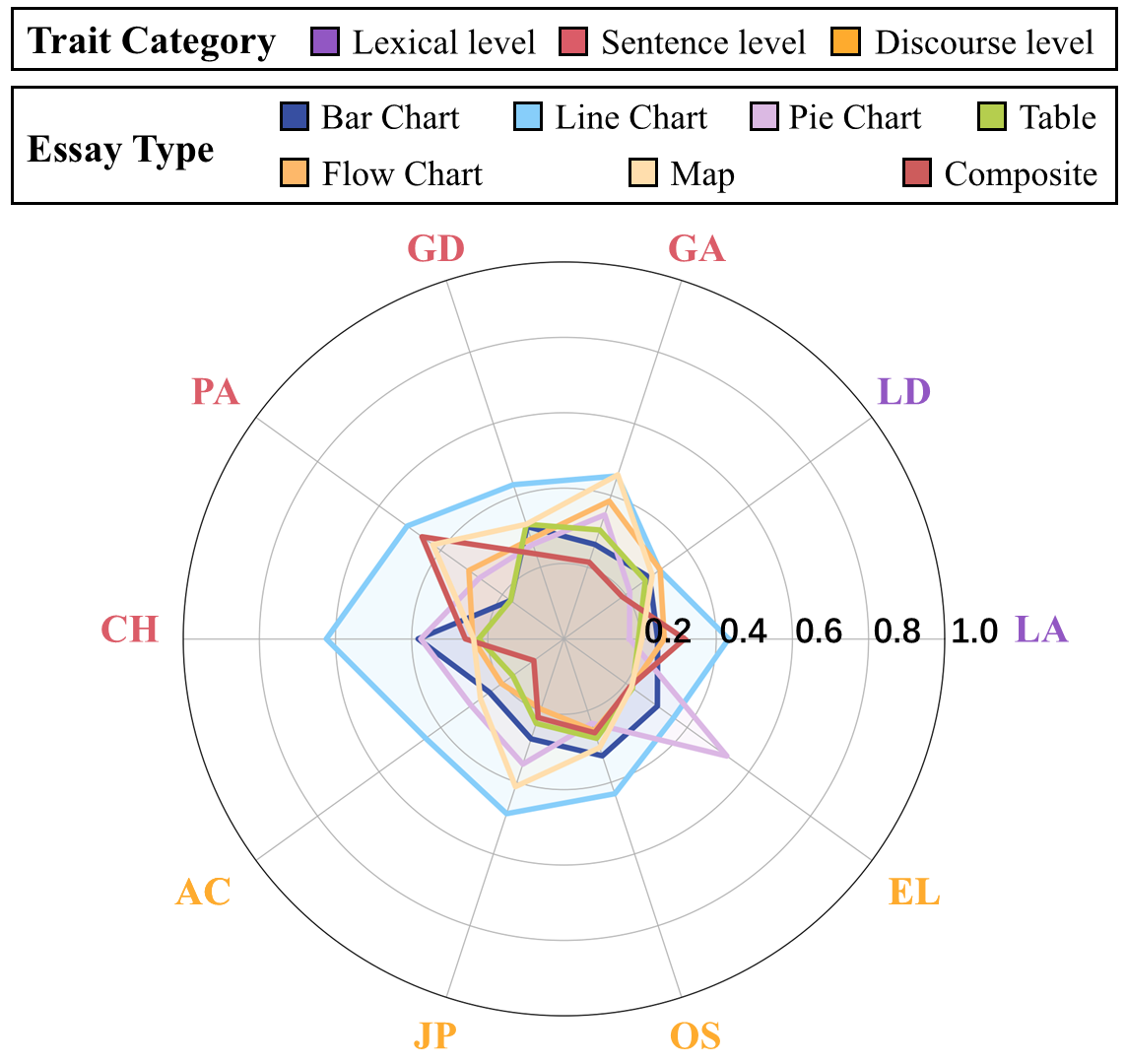}
  \caption{The relation between essay type and InternVL2's QWK.}
\label{fig:internvl2_type}
\end{figure*}

\begin{figure*}[htbp]
  \centering
  \includegraphics[width=0.5\textwidth]{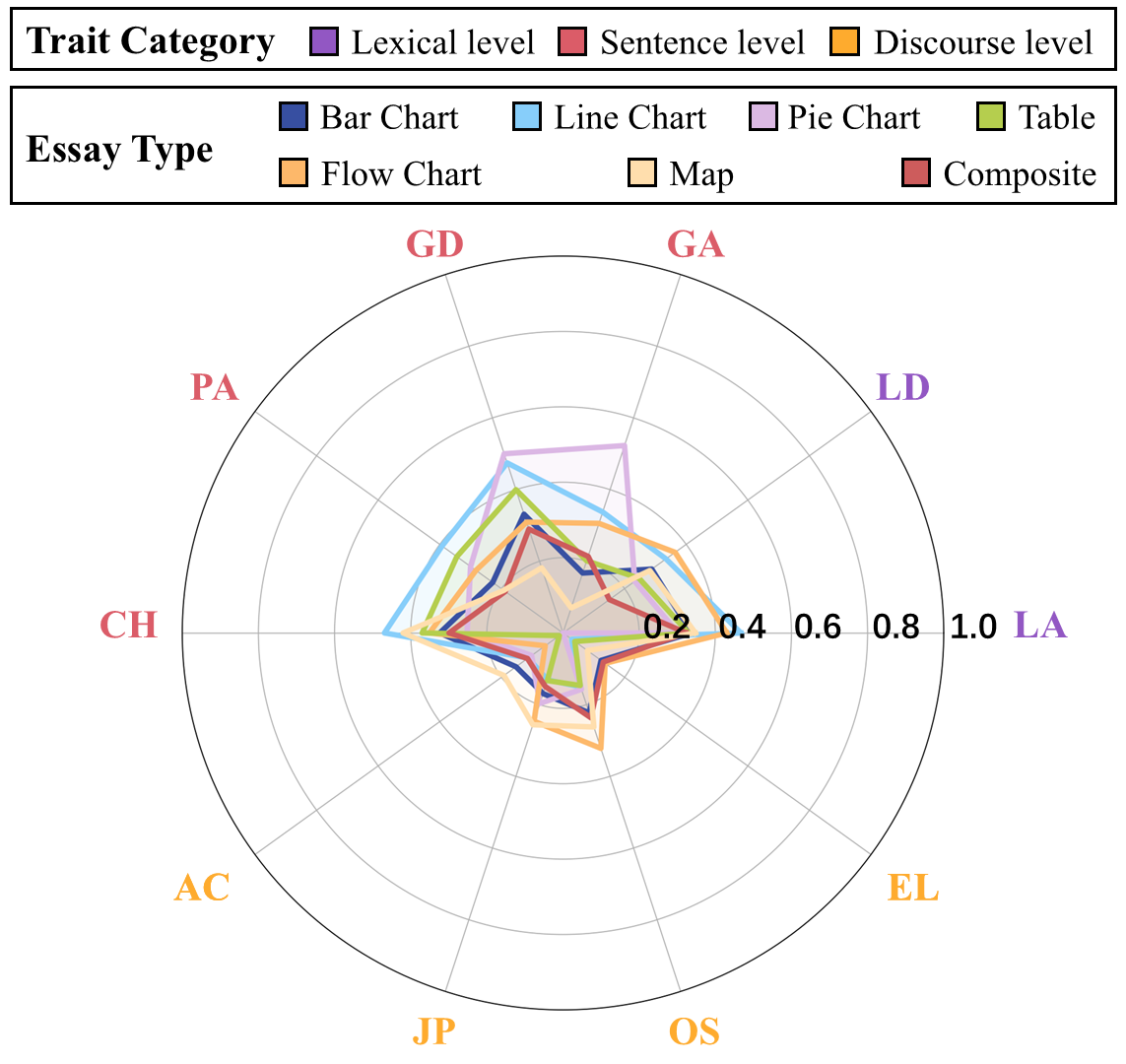}
  \caption{The relation between essay type and MiniCPM-LLaMA3-V2.5's QWK.}
\label{fig:minicpm_type}
\end{figure*}

\begin{figure*}[htbp]
  \centering
  \includegraphics[width=0.5\textwidth]{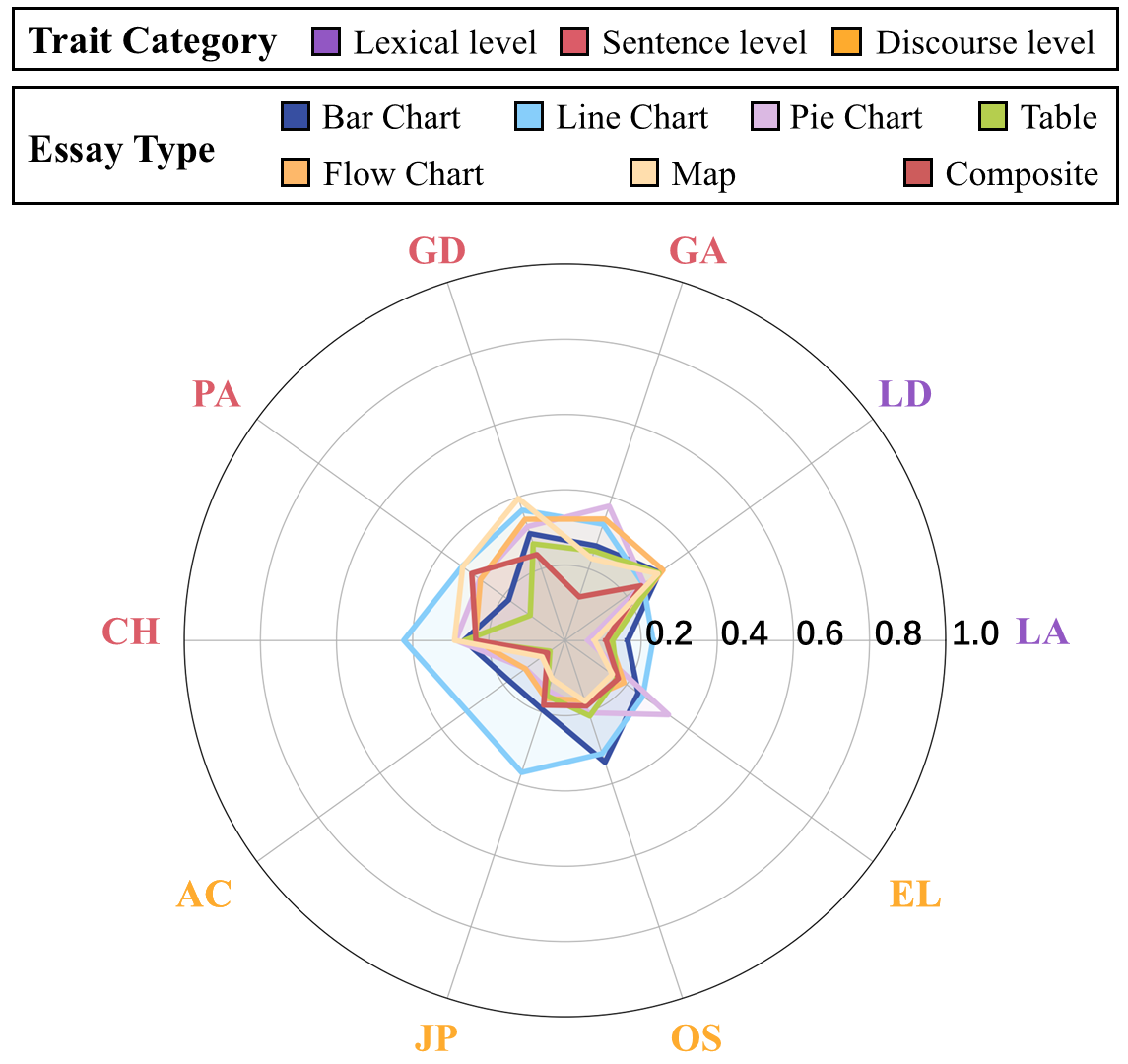}
  \caption{The relation between essay type and InternVL2.5's QWK.}
\label{fig:internvl2.5_type}
\end{figure*}

 We present the performance results of the top three open-source and closed-source MLLMs in grading essays related to line charts. Aside from the best performer, GPT-4o, other models evaluated include Claude-3.5-Sonnet, GPT-4o-mini, InternVL2, MiniCPM-LLaMA3-V2.5, and InternVL2.5. As shown in Figure \ref{fig:claude_sonnet_type} to Figure \ref{fig:internvl2.5_type} These models also demonstrated outstanding performance in assessing coherence when grading essays related to line charts.

\section{More on Analysis of \#image}
\label{app:single_multi}
\begin{figure*}[htbp]
  \centering
  \includegraphics[width=0.5\textwidth]{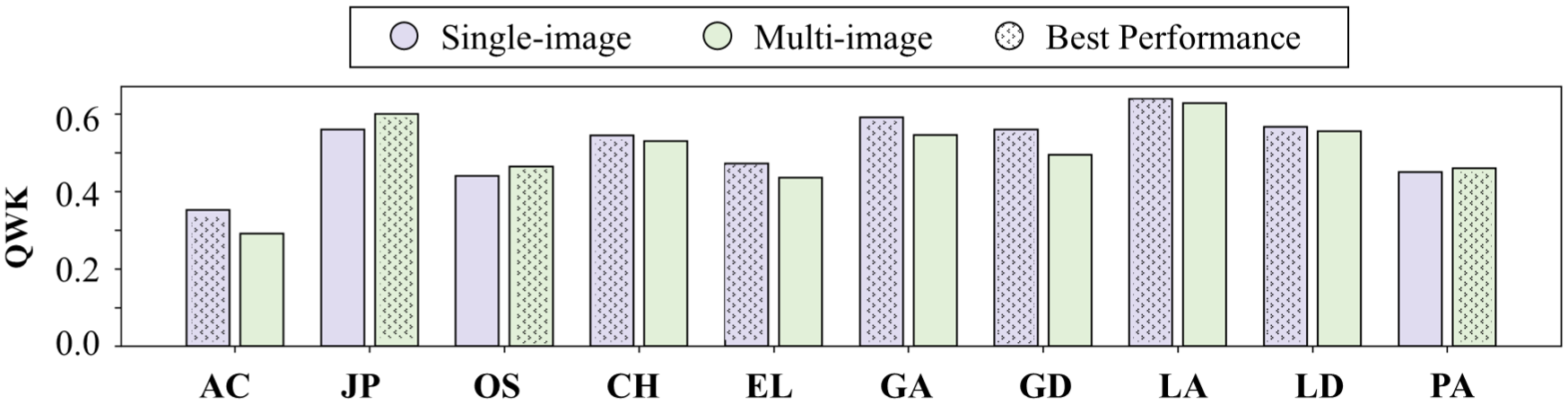}
  \caption{GPT-4o-mini’s QWK values across traits for single-image and multi-image settings.}
\label{fig:gpt40mini_single_multi}
\end{figure*}

\begin{figure*}[htbp]
  \centering
  \includegraphics[width=0.5\textwidth]{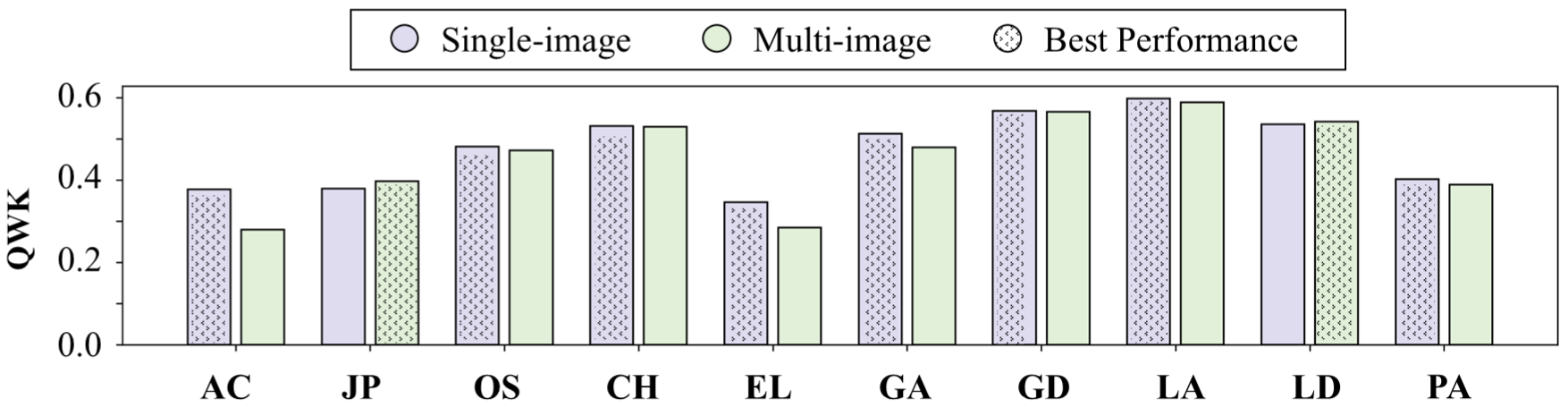}
  \caption{Claude 3.5 Haiku’s QWK values across traits for single-image and multi-image settings.}
\label{fig:claude_haiku_single_multi}
\end{figure*}

\begin{figure*}[htbp]
  \centering
  \includegraphics[width=0.5\textwidth]{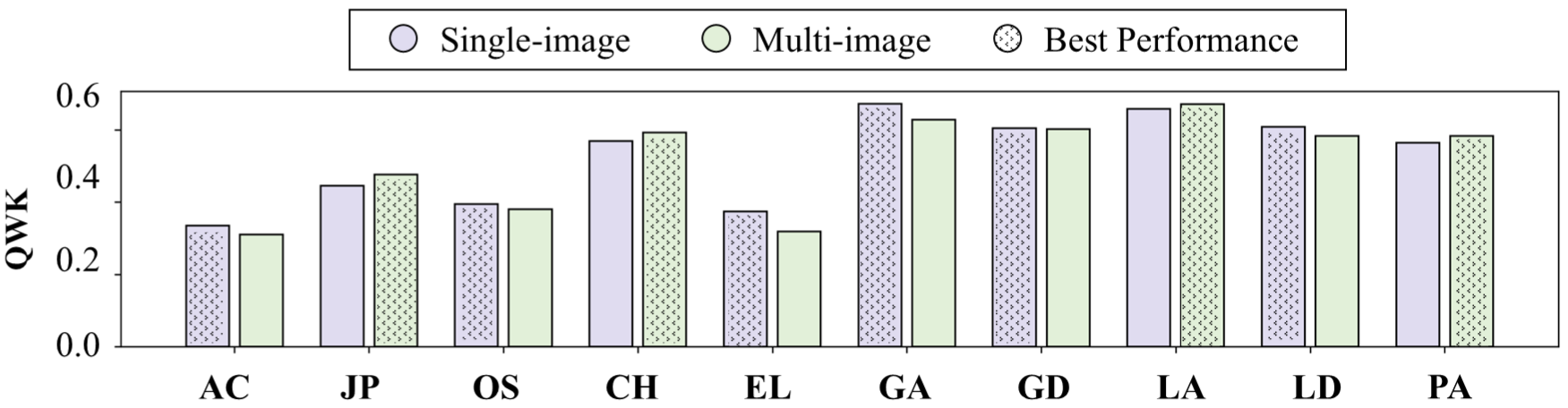}
  \caption{Claude 3.5 Sonnet’s QWK values across traits for single-image and multi-image settings.}
\label{fig:claude_sonnet_single_multi}
\end{figure*}

\begin{figure*}[htbp]
  \centering
  \includegraphics[width=0.5\textwidth]{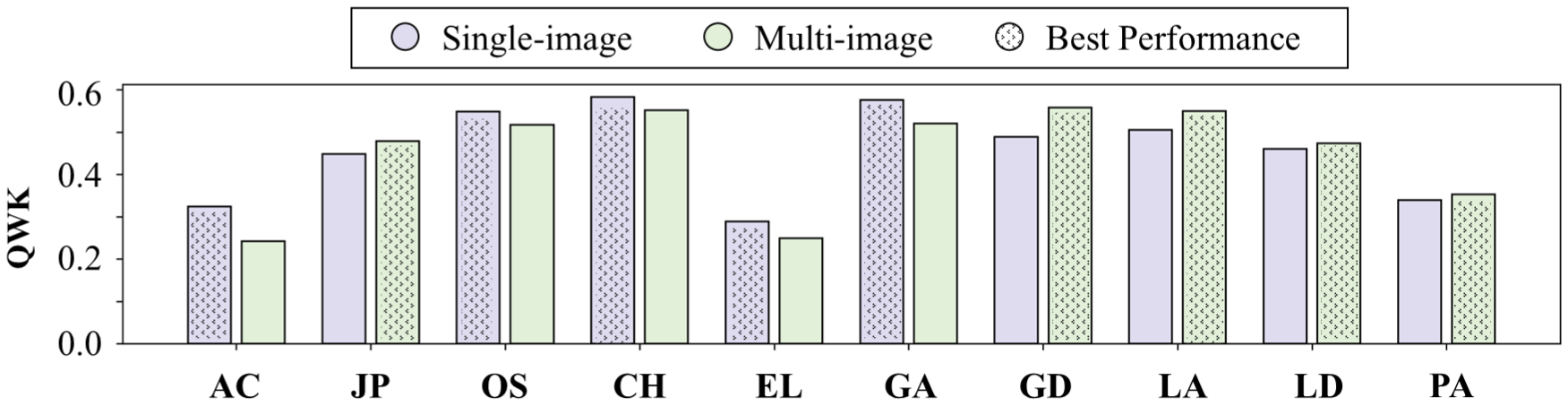}
  \caption{Gemini 1.5 Pro’s QWK values across traits for single-image and multi-image settings.}
\label{fig:Gemini_1_5_Pro_single_multi}
\end{figure*}

\begin{figure*}[htbp]
  \centering
  \includegraphics[width=0.5\textwidth]{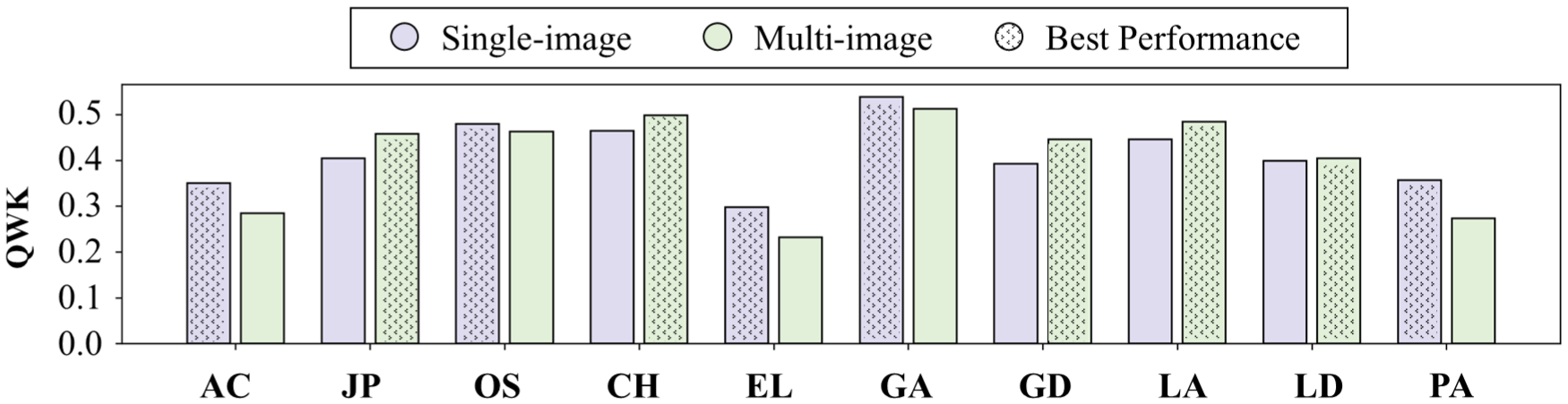}
  \caption{Gemini 1.5 Flash’s QWK values across traits for single-image and multi-image settings.}
\label{fig:Gemini_1_5_flash_single_multi}
\end{figure*}

\begin{figure*}[htbp]
  \centering
  \includegraphics[width=0.5\textwidth]{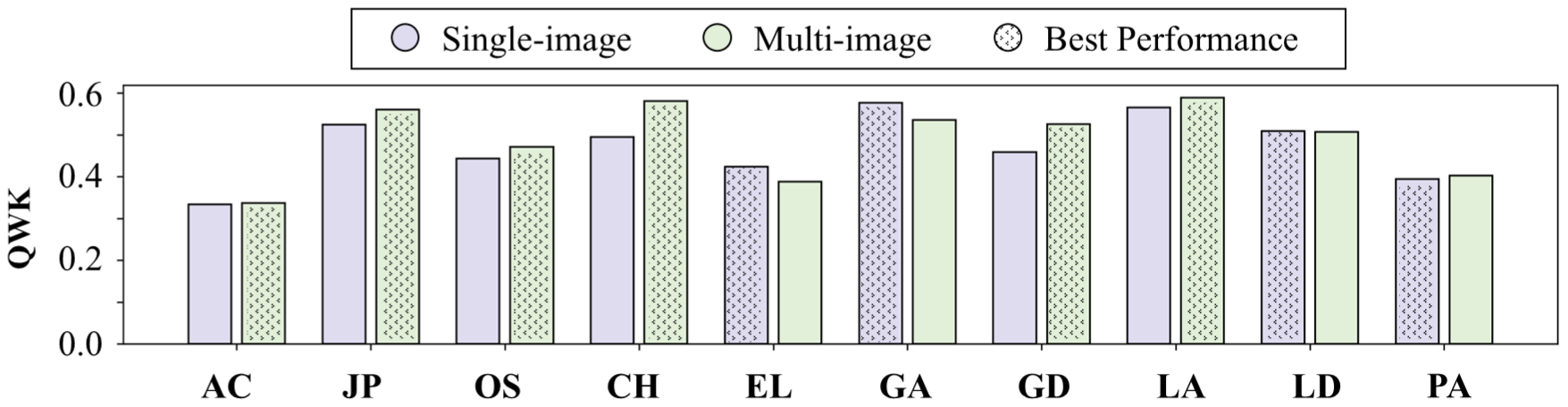}
  \caption{Qwen-Max’s QWK values across traits for single-image and multi-image settings.}
\label{fig:qwen_max_single_multi}
\end{figure*}

\begin{figure*}[htbp]
  \centering
  \includegraphics[width=0.5\textwidth]{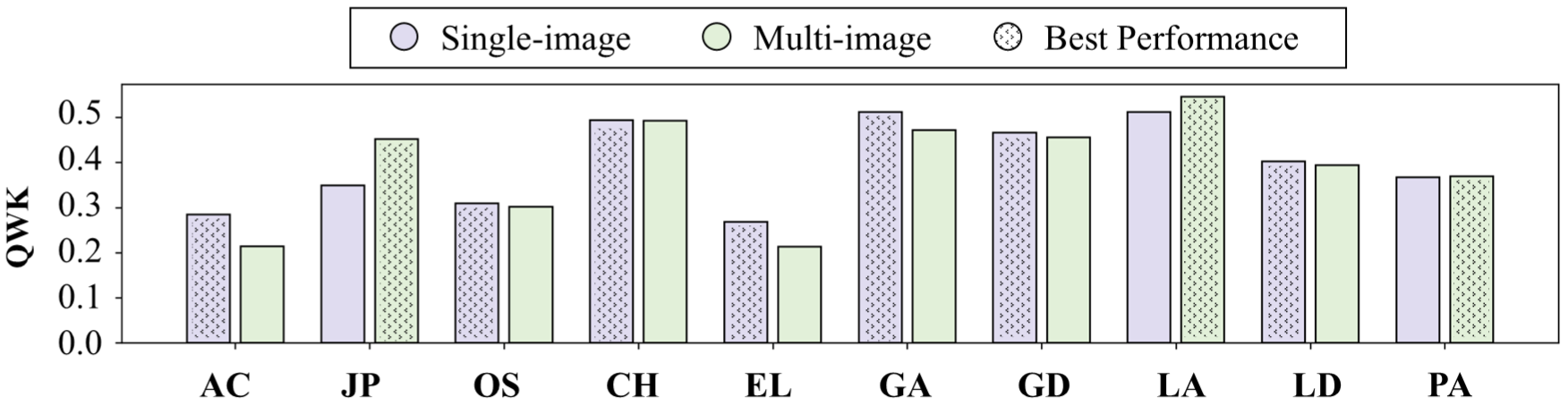}
  \caption{Step-1V’s QWK values across traits for single-image and multi-image settings.}
\label{fig:step1v_single_multi}
\end{figure*}

This section presents the evaluation results of all assessed closed-source MLLMs. As shown in Figure \ref{fig:gpt40mini_single_multi} to Figure \ref{fig:step1v_single_multi}, most closed-source MLLMs, except for the Gemini series and Qwen-Max, perform better when grading essays based on single-image tasks compared to multi-image tasks.
\section{More Multimodal Essay Scoring Examples}
\label{app:example}
\begin{figure*}[!t]
\vspace{-4mm}
  \centering
  \includegraphics[width=\textwidth]{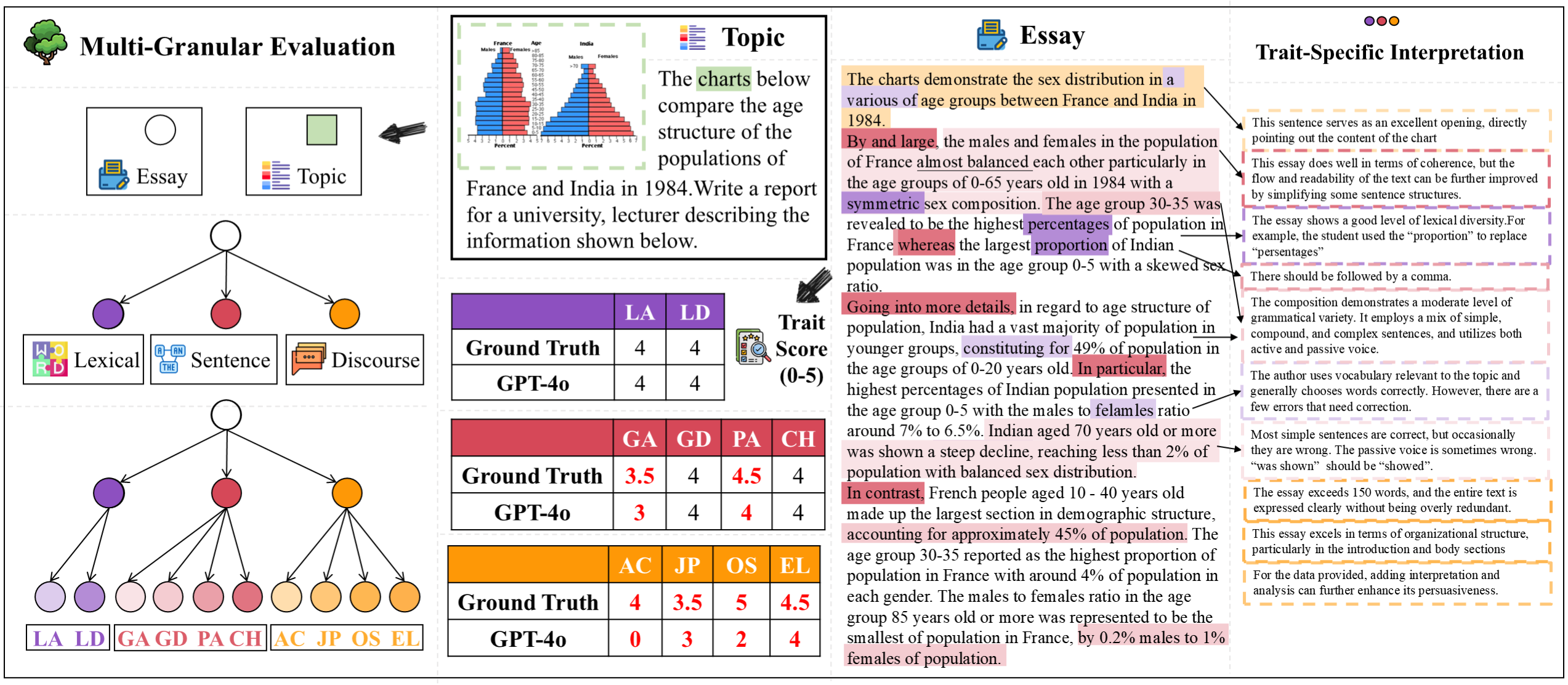}
  \caption{The example one of multi-granular evaluation for the essay.}
\label{fig:case_study_app1}
\vspace{-6mm}
\end{figure*}

\begin{figure*}[!t]
\vspace{-4mm}
  \centering
  \includegraphics[width=\textwidth]{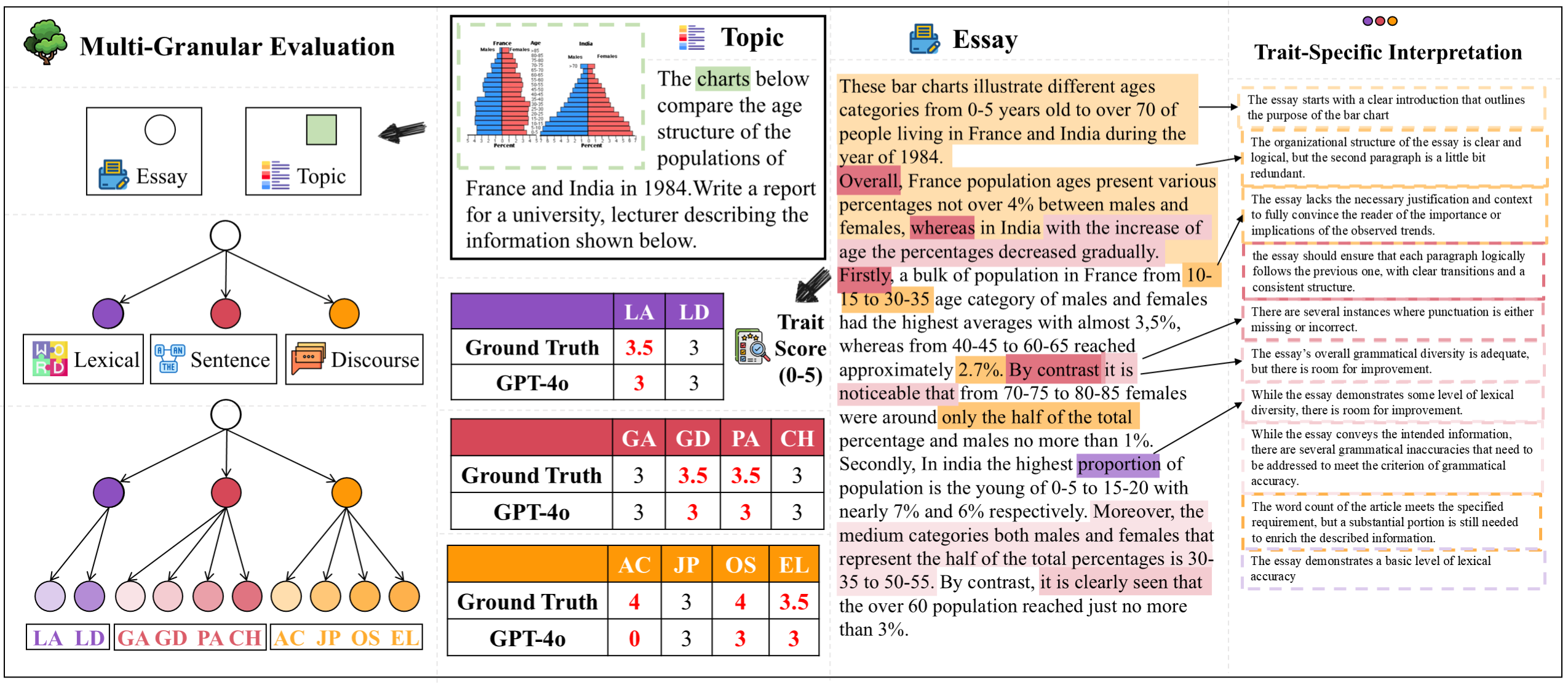}
  \caption{The example two of multi-granular evaluation for the essay.}
\label{fig:case_study_app2}
\vspace{-6mm}
\end{figure*}

\begin{figure*}[!t]
\vspace{-4mm}
  \centering
  \includegraphics[width=\textwidth]{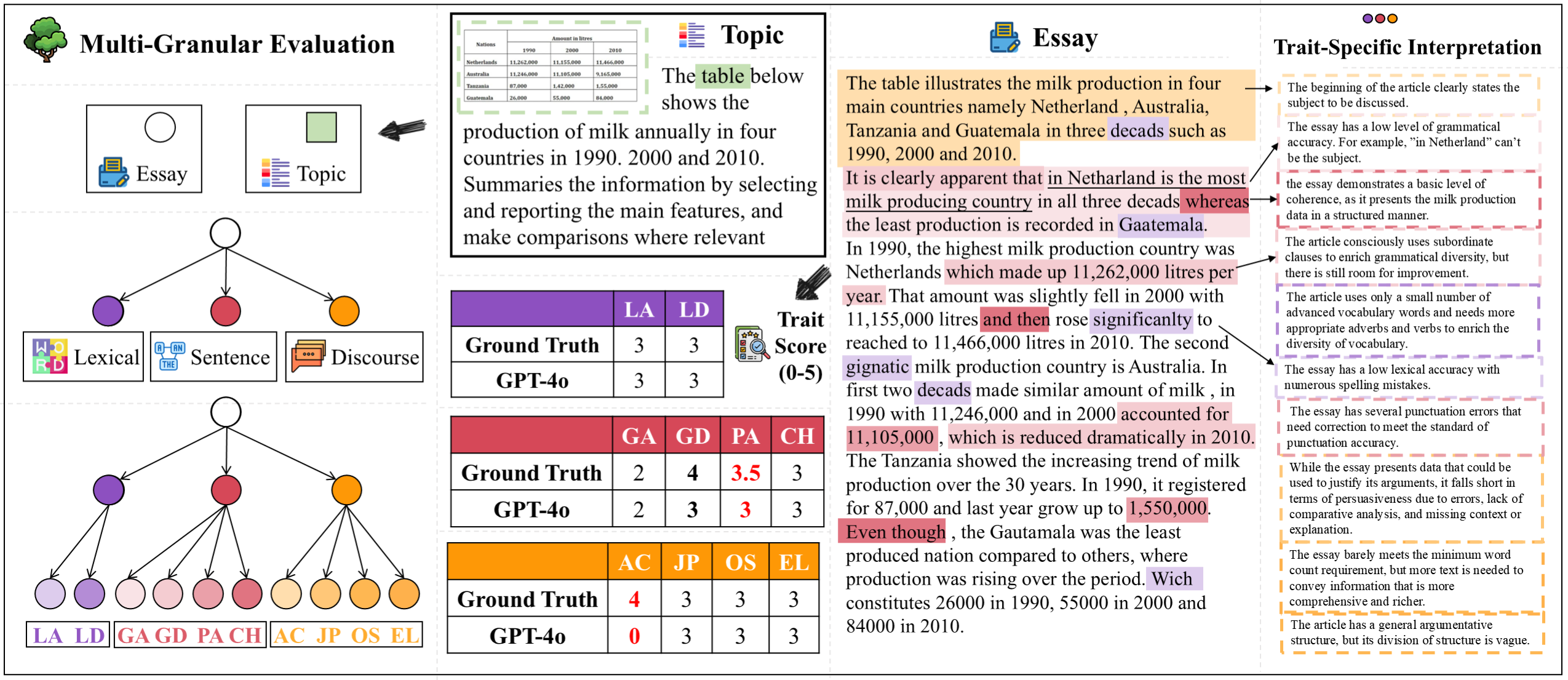}
  \caption{The example three of multi-granular evaluation for the essay.}
\label{fig:case_study_app3}
\vspace{-6mm}
\end{figure*}

This section provides additional examples of multimodal essay scoring based on Multi-Granular rubrics, which is shown in Figure \ref{fig:case_study_app1} to Figure \ref{fig:case_study_app3}, showcasing the application of our proposed framework to a diverse set of essays that incorporate both textual and visual elements. 

\clearpage

\begin{table*}[htbp]
  \centering
  \begin{tabularx}{\textwidth}{>{\centering\arraybackslash}p{0.7cm}>{\raggedright\arraybackslash}X} 
    \toprule
    \multicolumn{1}{c}{\textbf{Score}} & \multicolumn{1}{c}{\textbf{Scoring Criteria}} \\ 
    \midrule
    5 & The central argument is clear, and the first paragraph clearly outlines the topic of the image and question, providing guidance with no ambiguity.   \\ 
    \midrule    
    4 & The central argument is clear, and the first paragraph mentions the topic of the image and question, but the guidance is slightly lacking or the expression is somewhat vague.  \\ 
    \midrule    
    3 & The argument is generally clear, but the expression is vague, and it doesn't adequately guide the rest of the essay. \\ 
    \midrule    
    2 & The argument is unclear, the description is vague or incomplete, and it doesn't guide the essay.   \\ 
    \midrule    
    1 & The argument is vague, and the first paragraph fails to effectively summarize the topic of the image or question.    \\ 
    \midrule    
    0 & No central argument is presented, or the essay completely deviates from the topic and image.    \\ 
    \bottomrule
  \end{tabularx}
  \caption{Rubrics for evaluating the argument clarity of the essays.}
  \label{tab:AC}
\end{table*}

\begin{table*}[htbp]
  \centering
  \begin{tabularx}{\textwidth}{>{\centering\arraybackslash}p{0.7cm}>{\raggedright\arraybackslash}X} 
    \toprule
    \multicolumn{1}{c}{\textbf{Score}} & \multicolumn{1}{c}{\textbf{Scoring Criteria}} \\ 
    \midrule
    5  & Transitions between sentences are natural, and logical connections flow smoothly; appropriate use of linking words and transitional phrases.   \\ 
    \midrule
    4  & Sentences are generally coherent, with some transitions slightly awkward; linking words are used sparingly but are generally appropriate.   \\ 
    \midrule    
    3  & The logical connection between sentences is not smooth, with some sentences jumping or lacking flow; linking words are used insufficiently or inappropriately. \\ 
    \midrule    
    2  & Logical connections are weak, sentence connections are awkward, and linking words are either used too little or excessively.    \\ 
    \midrule    
    1  & There is almost no logical connection between sentences, transitions are unnatural, and linking words are very limited or incorrect.     \\ 
    \midrule    
    0  & No coherence at all, with logical confusion between sentences.    \\ 
    \bottomrule
  \end{tabularx}
  \caption{Rubrics for evaluating the coherence of the essays.}
  \label{tab:CH}
\end{table*}

\begin{table*}[htbp]
  \centering
  \begin{tabularx}{\textwidth}{>{\centering\arraybackslash}p{0.7cm}>{\raggedright\arraybackslash}X} 
    \toprule
    \multicolumn{1}{c}{\textbf{Score}} & \multicolumn{1}{c}{\textbf{Scoring Criteria}} \\ 
    \midrule
    5 & Word count is 150 words or more, with the content being substantial and without obvious excess or brevity.    \\ 
    \midrule
    4 & Word count is around 150 words, but slightly off (within 10 words), and the content is complete.    \\ 
    \midrule
    3 & Word count is noticeably too short or too long, and the content is not sufficiently substantial or is somewhat lengthy.  \\ 
    \midrule
    2 & Word count deviates significantly, failing to fully cover the requirements of the prompt.    \\ 
    \midrule
    1 & Word count is far below the requirement, and the content is incomplete.     \\ 
    \midrule
    0 & Word count is severely insufficient or excessive, making it impossible to meet the requirements of the prompt.    \\ 
    \bottomrule
  \end{tabularx}
  \caption{Rubrics for evaluating the essay length of the essays.}
  \label{tab:EL}
\end{table*}

\begin{table*}[htbp]
  \centering
  \begin{tabularx}{\textwidth}{>{\centering\arraybackslash}p{0.7cm}>{\raggedright\arraybackslash}X} 
    \toprule
    \multicolumn{1}{c}{\textbf{Score}} & \multicolumn{1}{c}{\textbf{Scoring Criteria}} \\ 
    \midrule
    5 & Sentence structure is accurate with no grammatical errors; both simple and complex sentences are error-free.    \\    
    \midrule
    4 & Sentence structure is generally accurate, with occasional minor errors that do not affect understanding; some errors in complex sentence structures.   \\    
    \midrule
    3 & Few grammatical errors, but more noticeable errors that affect understanding; simple sentences are accurate, but complex sentences frequently contain errors.  \\   
    \midrule
    2 & Numerous grammatical errors, with sentence structure affecting understanding; simple sentences are occasionally correct, but complex sentences have frequent errors.    \\  
    \midrule
    1 & A large number of grammatical errors, with sentence structure severely affecting understanding; sentence structure is unstable, and even simple sentences contain mistakes.     \\  
    \midrule
    0 & Sentence structure is completely incorrect, nonsensical, and difficult to understand.    \\ 
    \bottomrule
  \end{tabularx}
  \caption{Rubrics for evaluating the grammatical accuracy of the essays.}
  \label{tab:GA}
\end{table*}

\begin{table*}[htbp]
  \centering
  \begin{tabularx}{\textwidth}{>{\centering\arraybackslash}p{0.7cm}>{\raggedright\arraybackslash}X} 
    \toprule
    \multicolumn{1}{c}{\textbf{Score}} & \multicolumn{1}{c}{\textbf{Scoring Criteria}} \\ 
    \midrule
    5  & Uses a variety of sentence structures, including both simple and complex sentences, with flexible use of clauses and compound sentences, demonstrating rich sentence variation.     \\   
    \midrule
    4  & Generally uses a variety of sentence structures, with appropriate use of common clauses and compound sentences. Sentence structures vary, though some sentence types lack flexibility.    \\  
    \midrule
    3  & Uses a variety of sentence structures, but with limited use of complex sentences, which often contain errors. Sentence variation is somewhat restricted.  \\     
    \midrule
    2  & Sentence structures are simple, primarily relying on simple sentences, with occasional attempts at complex sentences, though errors occur frequently.    \\     
    \midrule
    1  & Sentence structures are very basic, with almost no complex sentences, and even simple sentences contain errors.     \\  
    \midrule
    0  & Only uses simple, repetitive sentences with no complex sentences, resulting in rigid sentence structures.    \\ 
    \bottomrule
  \end{tabularx}
  \caption{Rubrics for evaluating the grammatical diversity of the essays.}
  \label{tab:GD}
\end{table*}

\begin{table*}[htbp]
  \centering
  \begin{tabularx}{\textwidth}{>{\centering\arraybackslash}p{0.7cm}>{\raggedright\arraybackslash}X} 
    \toprule
    \multicolumn{1}{c}{\textbf{Score}} & \multicolumn{1}{c}{\textbf{Scoring Criteria}} \\ 
    \midrule
    5  & Fully addresses and accurately analyzes all important information in the image and prompt (\textit{e.g.}, data turning points, trends); argumentation is in-depth and logically sound.      \\    
    \midrule
    4  & Addresses most of the important information in the image and prompt, with reasonable analysis but slight shortcomings; argumentation is generally logical.     \\     
    \midrule
    3  &Addresses some important information in the image and prompt, but analysis is insufficient; argumentation is somewhat weak. \\     
    \midrule
    2  & Mentions a small amount of important information in the image and prompt, with simple or incorrect analysis; there are significant logical issues in the argumentation.   \\  
    \midrule
    1  & Only briefly mentions important information in the image and prompt or makes clear analytical errors, lacking reasonable reasoning.      \\   
    \midrule
    0  & Fails to mention key information from the image and prompt, lacks any argumentation, and is logically incoherent.   \\ 
    \bottomrule
  \end{tabularx}
  \caption{Rubrics for evaluating the justifying persuasiveness of the essays.}
  \label{tab:JP}
\end{table*}

\begin{table*}[htbp]
  \centering
  \begin{tabularx}{\textwidth}{>{\centering\arraybackslash}p{0.7cm}>{\raggedright\arraybackslash}X} 
    \toprule
    \multicolumn{1}{c}{\textbf{Score}} & \multicolumn{1}{c}{\textbf{Scoring Criteria}} \\ 
    \midrule
    5 & Vocabulary is accurately chosen, with correct meanings and spelling, and minimal errors; words are used precisely to convey the intended meaning.     \\ 
    \midrule
    4 & Vocabulary is generally accurate, with occasional slight meaning errors or minor spelling mistakes, but they do not affect overall understanding; words are fairly precise.     \\ 
    \midrule
    3 & Vocabulary is mostly correct, but frequent minor errors or spelling mistakes affect some expressions; word choice is not fully precise.  \\ 
    \midrule
    2 & Vocabulary is inaccurate, with significant meaning errors and frequent spelling mistakes, affecting understanding.    \\ 
    \midrule
    1 & Vocabulary is severely incorrect, with unclear meanings and noticeable spelling errors, making comprehension difficult.       \\ 
    \midrule
    0 & Vocabulary choice and spelling are completely incorrect, and the intended meaning is unclear or impossible to understand.   \\  
    \bottomrule
  \end{tabularx}
  \caption{Rubrics for evaluating the lexical accuracy of the essays.}
  \label{tab:LA}
\end{table*}

\begin{table*}[htbp]
  \centering
  \begin{tabularx}{\textwidth}{>{\centering\arraybackslash}p{0.7cm}>{\raggedright\arraybackslash}X} 
    \toprule
    \multicolumn{1}{c}{\textbf{Score}} & \multicolumn{1}{c}{\textbf{Scoring Criteria}} \\ 
    \midrule
    5  & Vocabulary is rich and diverse, with a wide range of words used flexibly, avoiding repetition.   \\ 
    \midrule
    4  & Vocabulary diversity is good, with a broad range of word choices, occasional repetition, but overall flexible expression.     \\ 
    \midrule
    3  &Vocabulary diversity is average, with some variety in word choice but limited, with frequent repetition.  \\ 
    \midrule
    2  & Vocabulary is fairly limited, with a lot of repetition and restricted word choice.   \\ 
    \midrule
    1  & Vocabulary is very limited, with frequent repetition and an extremely narrow range of words.       \\ 
    \midrule
    0  & Vocabulary is monotonous, with almost no variation, failing to demonstrate vocabulary diversity.  \\  
    \bottomrule
  \end{tabularx}
  \caption{Rubrics for evaluating the lexical diversity of the essays.}
  \label{tab:LD}
\end{table*}

\begin{table*}[htbp]
  \centering
  \begin{tabularx}{\textwidth}{>{\centering\arraybackslash}p{0.7cm}>{\raggedright\arraybackslash}X} 
    \toprule
    \multicolumn{1}{c}{\textbf{Score}} & \multicolumn{1}{c}{\textbf{Scoring Criteria}} \\ 
    \midrule
    5 & The essay has a well-organized structure, with clear paragraph divisions, each focused on a single theme. There are clear topic sentences and concluding sentences, and transitions between paragraphs are natural.  \\ 
    \midrule
    4 & The structure is generally reasonable, with fairly clear paragraph divisions, though transitions may be somewhat awkward and some paragraphs may lack clear topic sentences.      \\ 
    \midrule
    3 &The structure is somewhat disorganized, with unclear paragraph divisions, a lack of topic sentences, or weak logical flow.  \\ 
    \midrule
    2 & The structure is unclear, with improper paragraph divisions and poor logical coherence.    \\ 
    \midrule
    1 & The paragraph structure is chaotic, with most paragraphs lacking clear topic sentences and disorganized content.       \\ 
    \midrule
    0 & No paragraph structure, content is jumbled, and there is a complete lack of logical connections. \\
    \bottomrule
  \end{tabularx}
  \caption{Rubrics for evaluating the organizational structure of the essays.}
  \label{tab:OS}
\end{table*}

 \begin{table*}[htbp]
  \centering
  \begin{tabularx}{\textwidth}{>{\centering\arraybackslash}p{0.7cm}>{\raggedright\arraybackslash}X} 
    \toprule
    \multicolumn{1}{c}{\textbf{Score}} & \multicolumn{1}{c}{\textbf{Scoring Criteria}} \\ 
    \midrule
    5 & Punctuation is used correctly throughout, adhering to standard rules with no errors. \\ 
    \midrule
    4 & Punctuation is mostly correct, with occasional minor errors that do not affect understanding.     \\ 
    \midrule
    3 &Punctuation is generally correct, but there are some noticeable errors that slightly affect understanding.  \\ 
    \midrule
    2 & There are frequent punctuation errors, some of which affect understanding.    \\ 
    \midrule
    1 & Punctuation errors are severe, significantly affecting comprehension.       \\ 
    \midrule
    0 & Punctuation is completely incorrect or barely used, severely hindering understanding. \\ 
    \bottomrule
  \end{tabularx}
  \caption{Rubrics for evaluating the punctuation accuracy of the essays.}
  \label{tab:PA}
\end{table*}

\begin{table*}[htbp]
\centering
\small
\renewcommand\tabcolsep{5pt} 
\renewcommand\arraystretch{1.2} 
\begin{tabular}{c|p{0.5\textwidth}|c}
\hline
\textbf{Image} & \textbf{Topic} & \textbf{Frequency} \\
\hline
$\raisebox{-.78\height}{\includegraphics[width=0.2\textwidth]{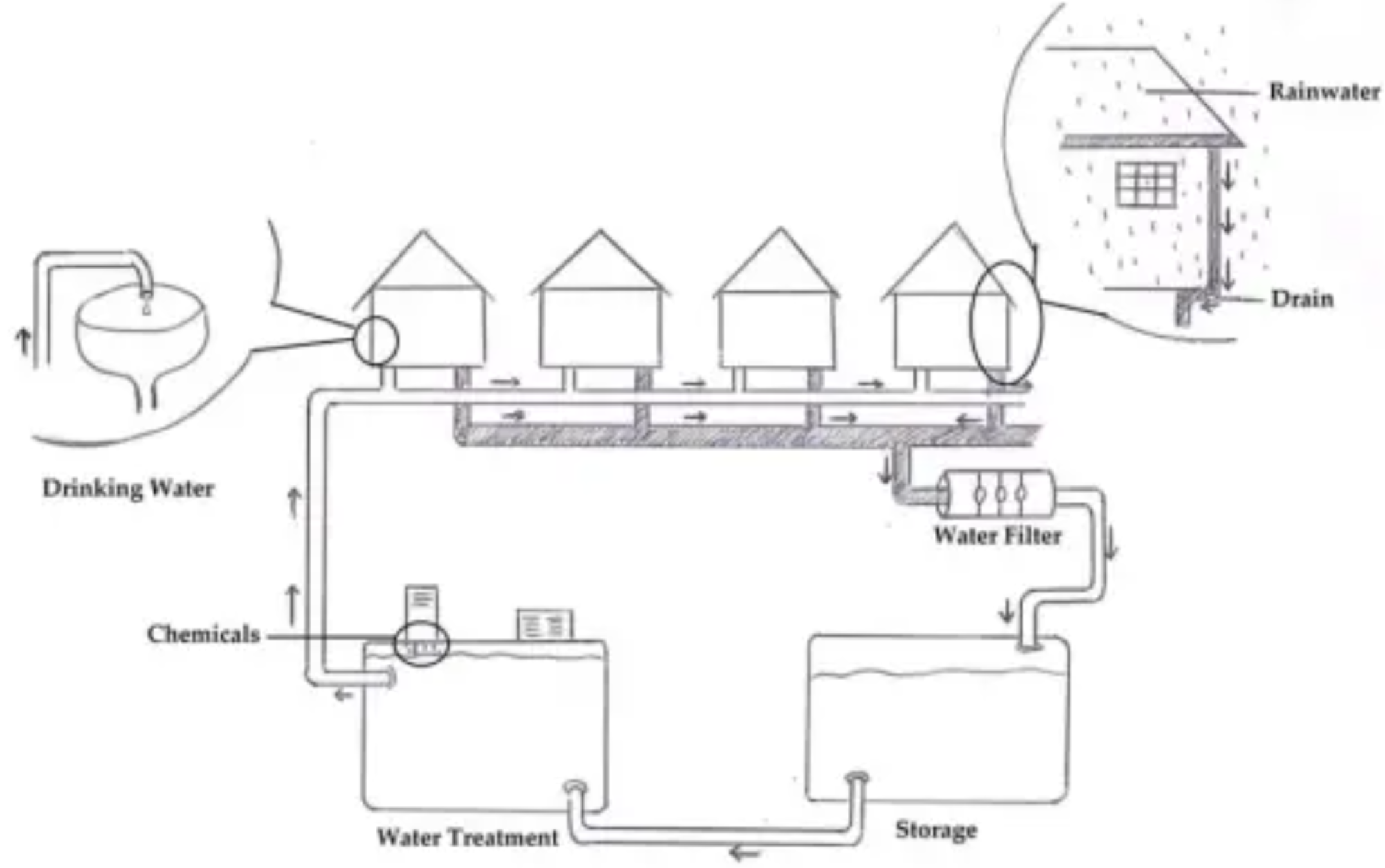}}$ & The diagram below shows how rain water is collected and then treated to be used as drinking water in an Australian town. Summarise the information by selecting and reporting the main features and make comparisons where relevant. You should write at least 150 words. & 23 \\
\hline
$\raisebox{-.72\height}{\includegraphics[width=0.2\textwidth]{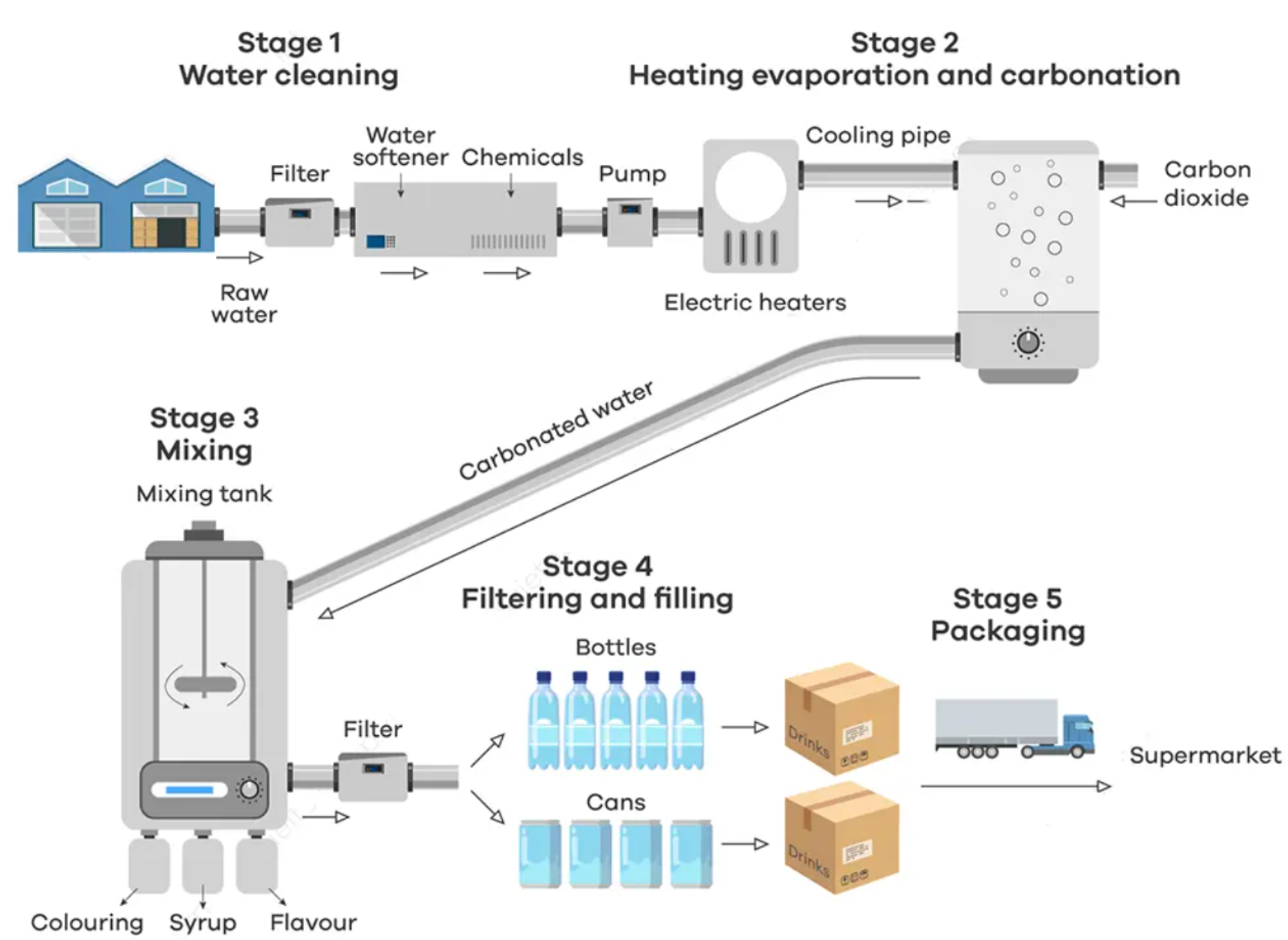}}$  & The diagram gives information about the process of making carbonated drinks. Summarise the information by selecting and report in the main features, and make comparisons where relevant. You should write at least 150 words. & 23 \\
\hline
$\raisebox{-.75\height}{\includegraphics[width=0.2\textwidth]{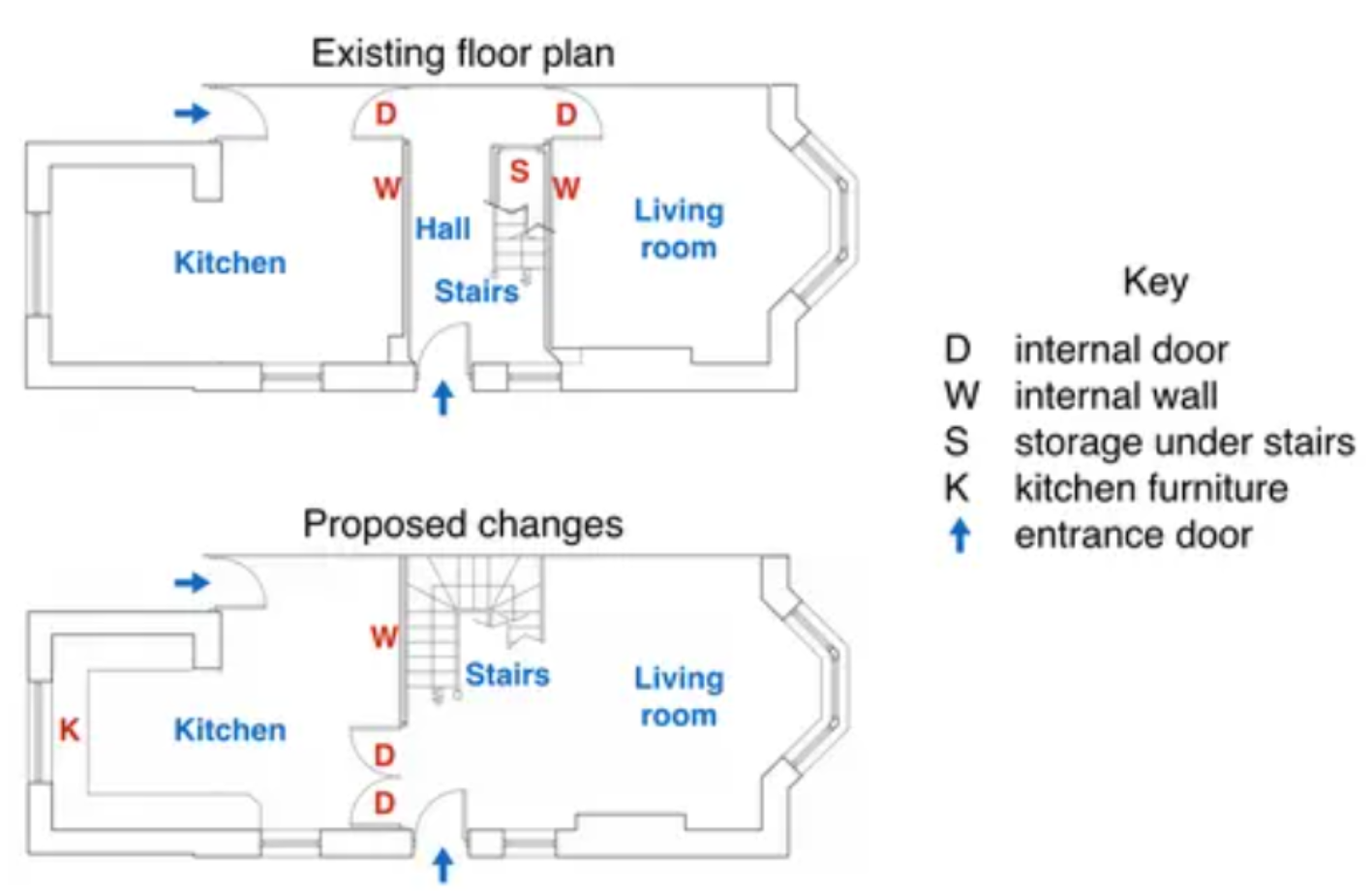}}$ & The diagrams below show the existing ground floor plan of a house and a proposed plan for some building work.Summarise the information by selecting and reporting the main features and make comparisons where relevant. You should write at least 150 words. & 16 \\
\hline
$\raisebox{-.75\height}{\includegraphics[width=0.2\textwidth]{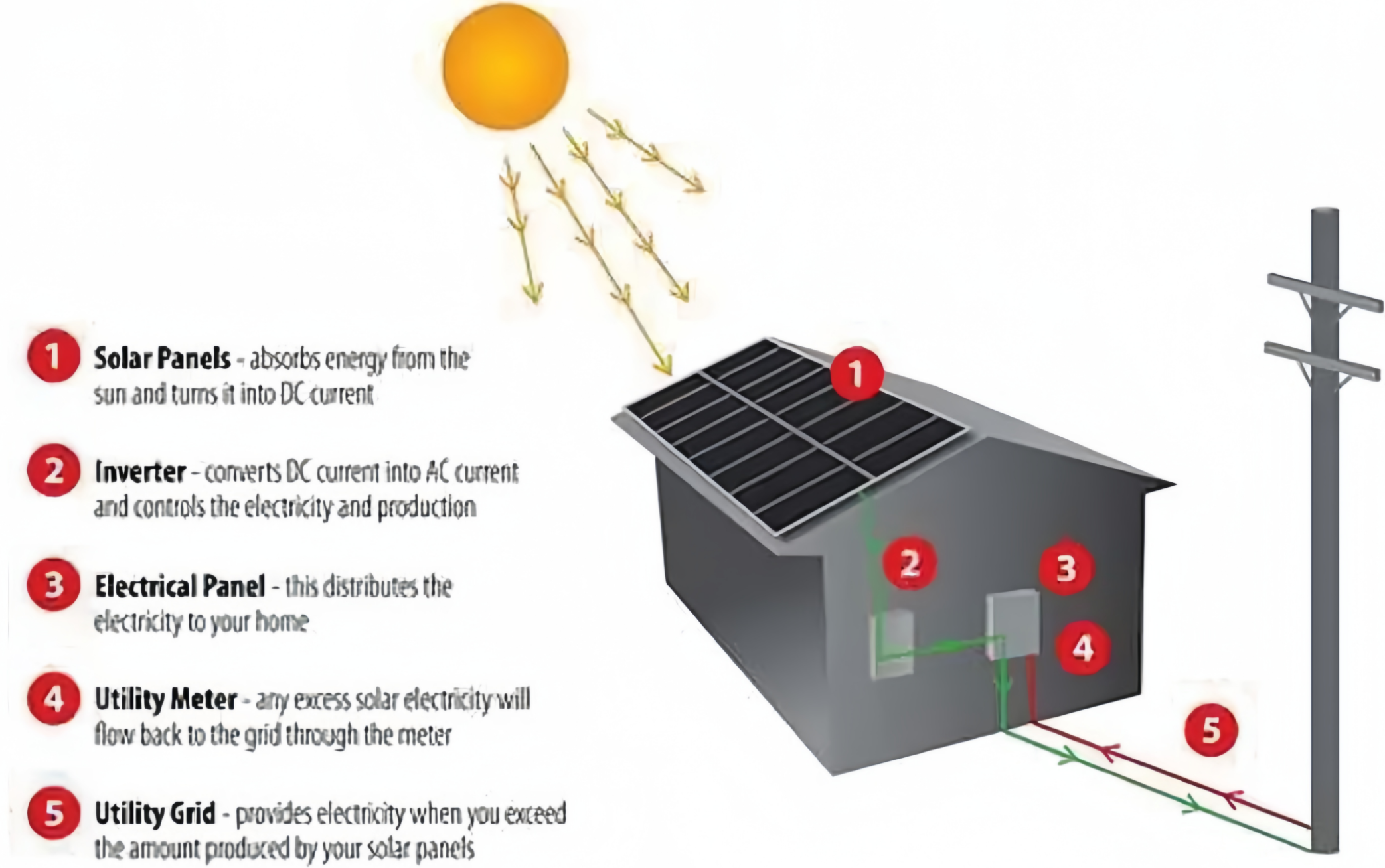}}$ & The diagram below shows how solar panels can be used to provide electricity for domestic use.Write a report for a university, lecturer describing the information shown below. You should write at least 150 words. & 16 \\
\hline
$\raisebox{-.75\height}{\includegraphics[width=0.2\textwidth]{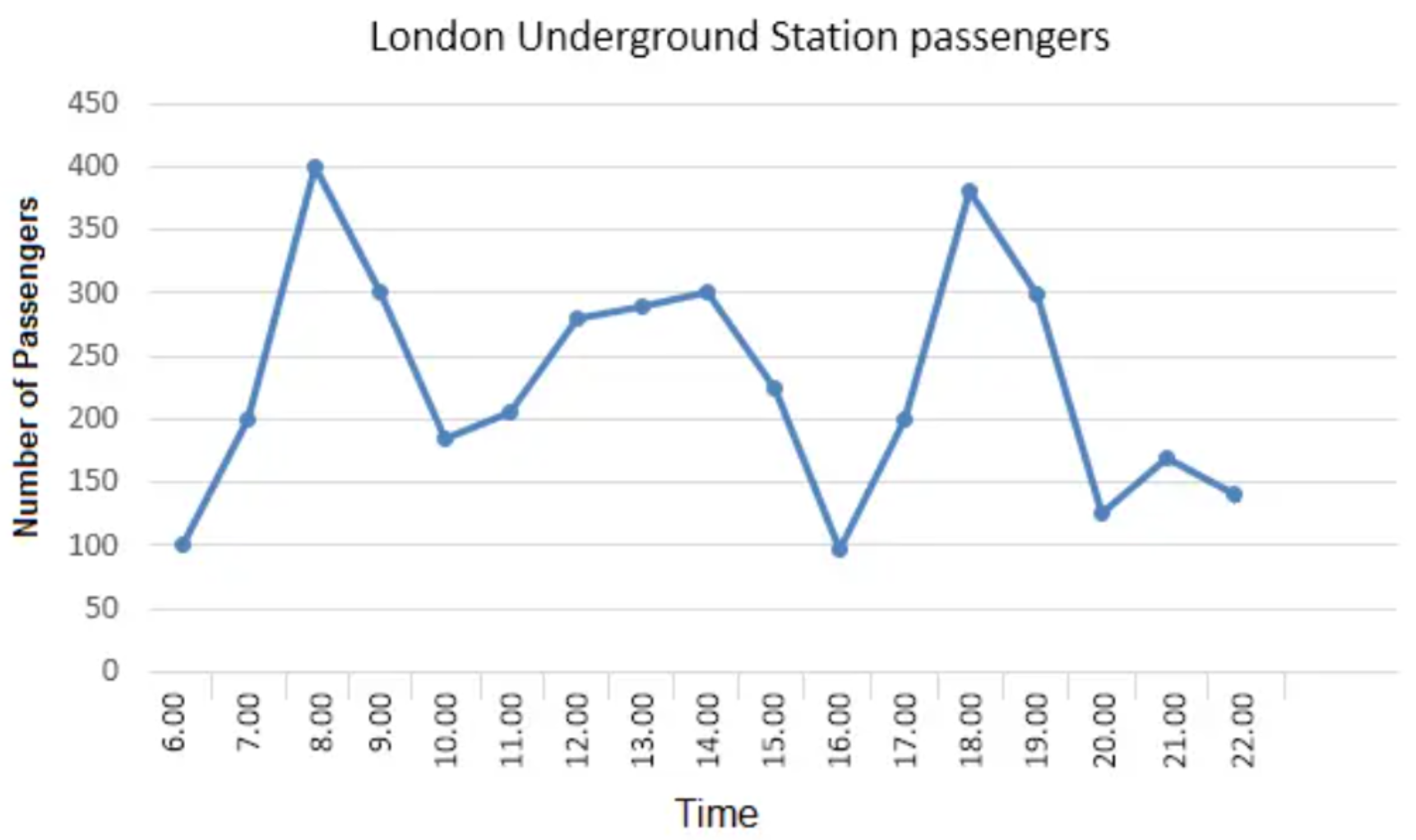}}$ & The graph shows Underground Station passenger numbers in London.Summarise the information by selecting and reporting the main features, and make comparisons where relevant. You should write at least 150 words. & 16 \\
\hline
\end{tabular}
\caption{The top five most frequent topics in the dataset with images, topics, and frequencies.}
\label{tab:topic}
\end{table*}

\begin{table*}[htbp]
\small
\centering
\begin{tabular}{p{0.3\linewidth} | p{0.25\linewidth} | p{0.25\linewidth}}
\toprule
\textbf{Trait} & \textbf{Qwen2-VL (7B)} & \textbf{Qwen2-VL (72B)}  \\
\midrule
Argument Clarity & 0.17 & 0.34 \\
\midrule
Justifying Persuasiveness & 0.10 & 0.52 \\
\midrule
Organizational Structure & 0.14 & 0.48 \\
\midrule
Coherence & 0.13 & 0.55 \\
\midrule
Essay Length & 0.15 & 0.40 \\
\midrule
Grammatical Accuracy & 0.21 & 0.51 \\
\midrule
Grammatical Diversity & 0.16 & 0.55 \\
\midrule
Lexical Accuracy & 0.20 & 0.52 \\
\midrule
Lexical Diversity & 0.26 & 0.38 \\
\midrule
Punctuation Accuracy & 0.12 & 0.42 \\
\bottomrule
\end{tabular}
\caption{QWK values of Qwen2-VL (7B) and Qwen2-VL (72B).}
\label{tab:scaling law}
\end{table*}

\clearpage

\begin{table*}[htbp]
\small
\centering
\begin{tabular}{p{0.3\linewidth} | p{0.1\linewidth} | p{0.2\linewidth} | p{0.1\linewidth}}
\toprule
\textbf{Trait} & \textbf{GPT-4o} & \textbf{GPT-4o-mini} & \textbf{GPT-3.5} \\
\midrule
Argument Clarity & 0.30 & 0.33 & 0.23 \\
\midrule
Justifying Persuasiveness & 0.8 & 0.57 & 0.24 \\
\midrule
Organizational Structure & 0.79 & 0.45 & 0.18 \\
\midrule
Coherence & 0.87 & 0.54 & 0.37\\
\midrule
Essay Length & 0.70 & 0.46 & 0.35 \\
\midrule
Grammatical Accuracy & 0.85 & 0.58 & 0.42 \\
\midrule
Grammatical Diversity & 0.61 & 0.54 & 0.43 \\
\midrule
Lexical Accuracy & 0.89 & 0.64 & 0.21 \\
\midrule
Lexical Diversity & 0.89 & 0.56 & 0.24  \\
\midrule
Punctuation Accuracy & 0.65 & 0.45 & 0.28 \\
\bottomrule
\end{tabular}
\caption{QWK values of GPT-4o, GPT-4o-mini and GPT-3.5 without image inputs.}
\label{tab:ablation study}
\end{table*}

\begin{figure*}[th!]
    \centering
    \begin{tcolorbox}[colback=white, colframe=black, enhanced jigsaw, listing only, listing options={basicstyle=\rmfamily}]
        \textbf{Task Definition:} Assume you are a professional English Educator. You need to score the \{Trait\} in the student's essay. Based on the essay topic and image prompt, as well as the student's essay, please assign a score (0-5) according to the criteria in the rubric. The output should be only the score. \\[1em]
        \textbf{Rubrics:} \{Trait-specific corresponding rubrics\}
        \\[1em]
        \textbf{Below is the reference content:} \\
        Image: "\{image\}"\\
        Essay Topic: "\{question\}"\\
        Student's Essay: "\{essay\}" \\[1em]
        \textbf{Instruction:} Please output only the number of the score (e.g. 5)
    \end{tcolorbox}
    \caption{Prompt for trait-specific AES task.}
    \label{fig:prompt example}
\end{figure*}

\begin{table*}[htbp]
\small
\centering
\begin{tabular}{p{0.25\linewidth} | p{0.25\linewidth} | p{0.3\linewidth}}
\toprule
\textbf{MLLMs} & \textbf{Source} & \textbf{URL} \\
\midrule
Yi-VL-6B & local checkpoint & \url{https://huggingface.co/01-ai/Yi-VL-6B} \\
\midrule
Qwen2-VL-7B & local checkpoint & \url{https://huggingface.co/Qwen/Qwen2-VL-7B} \\
\midrule
DeepSeek-VL-7B & local checkpoint & \url{https://huggingface.co/deepseek-ai/deepseek-vl-7b-chat} \\
\midrule
InternVL2-8B & local checkpoint &  \url{https://huggingface.co/OpenGVLab/InternVL2-8B}\\
\midrule
InternVL2.5-8B & local checkpoint & \url{https://huggingface.co/OpenGVLab/InternVL2_5-8B} \\
\midrule
MiniCPM-V 2.6-8B & local checkpoint & \url{https://huggingface.co/openbmb/MiniCPM-V-2_6} \\
\midrule
MiniCPM-Llama3-V 2.5-8B & local checkpoint & \url{https://huggingface.co/openbmb/MiniCPM-Llama3-V-2_5} \\
\midrule
LLaMA-3.2-Vision-Instruct-11B & local checkpoint & \url{https://huggingface.co/meta-llama/Llama-3.2-11B-Vision-Instruct} \\
\midrule
Qwen-Max & \texttt{qwen-vl-max-0809} & \url{https://modelscope.cn/studios/qwen/Qwen-VL-Max} \\
\midrule
Step-1V & \texttt{step-1v-32k} & \url{https://platform.stepfun.com/docs/llm/vision} \\
\midrule
Gemini 1.5 Pro & \texttt{gemini-1.5-pro-latest} & \url{https://deepmind.google/technologies/gemini/pro/} \\
\midrule
Gemini 1.5 Flash  & \texttt{gemini-1.5-flash-latest} & \url{https://ai.google.dev/gemini-api/docs/models/gemini#gemini-1.5-flash} \\
\midrule
Claude 3.5 Haiku & \texttt{claude-3.5-haiku-20241022} & \url{https://www.anthropic.com/claude/haiku} \\
\midrule
Claude 3.5 Sonnet & \texttt{claude-3.5-sonnet-20241022} & \url{https://www.anthropic.com/claude/sonnet} \\
\midrule
GPT-4o-mini & \texttt{gpt-4o-mini-2024-07-18} & \url{https://platform.openai.com/docs/models/gpt-4o-mini} \\
\midrule
GPT-4o & \texttt{gpt-4o-2024-08-06} & \url{https://platform.openai.com/docs/models/gpt-4o}\\
\bottomrule
\end{tabular}
\caption{Sources of our evaluated MLLMs.}
\label{tab:mllm hyperparams}
\end{table*}

\clearpage

\clearpage

\clearpage

\end{document}